\documentclass[10pt,twocolumn]{article}
\usepackage[left=1in,right=1in,top=0.85in,bottom=0.85in,columnsep=0.24in]{geometry}
\usepackage{amsmath,amssymb,bm}
\usepackage{graphicx}
\usepackage{array}
\usepackage{booktabs}
\usepackage{multirow}
\usepackage{xcolor}
\usepackage{algorithm}
\usepackage{algpseudocode}
\usepackage{tikz}
\usetikzlibrary{arrows.meta,positioning,shapes.geometric,fit}
\definecolor{lpInk}{HTML}{3F3833}
\definecolor{lpMuted}{HTML}{7A6F64}
\definecolor{lpLine}{HTML}{D8CEC1}
\definecolor{lpPaper}{HTML}{FCF9F4}
\definecolor{lpSand}{HTML}{E9DFD1}
\definecolor{lpSandEdge}{HTML}{A39584}
\definecolor{lpCross}{HTML}{6F655B}
\definecolor{lpAccent}{HTML}{B5532A}
\definecolor{lpAccentDark}{HTML}{8C3B1B}
\definecolor{lpAccentFill}{HTML}{D9774A}
\definecolor{lpClay}{HTML}{F8E4D8}
\definecolor{lpSage}{HTML}{E6ECDA}
\definecolor{lpSageEdge}{HTML}{6F7F4E}
\definecolor{lpNeutral}{HTML}{F4EFE7}
\usepackage[colorlinks=true,linkcolor=blue!60!black,citecolor=blue!60!black,urlcolor=blue!60!black]{hyperref}
\usepackage{microtype}
\usepackage{placeins}
\usepackage{float}
\usepackage{balance}
\usepackage[font=small,labelfont=bf,skip=5pt]{caption}

\makeatletter
\renewcommand\section{\@startsection{section}{1}{\z@}%
  {-3.1ex \@plus -1ex \@minus -.2ex}{2.0ex \@plus.2ex}%
  {\raggedright\normalfont\Large\bfseries}}
\renewcommand\subsection{\@startsection{subsection}{2}{\z@}%
  {-2.9ex\@plus -1ex \@minus -.2ex}{1.3ex \@plus .2ex}%
  {\raggedright\normalfont\large\bfseries}}
\renewcommand\subsubsection{\@startsection{subsubsection}{3}{\z@}%
  {-3.25ex\@plus -1ex \@minus -.2ex}{1.5ex \@plus .2ex}%
  {\raggedright\normalfont\normalsize\bfseries}}
\makeatother

\newcolumntype{P}[1]{>{\raggedright\arraybackslash}p{#1}}

\newcommand{\rpun}{resolved-positive\slash unresolved\slash resolved-negative}
\newcommand{\ours}{LP-BTS}
\newcommand{\ckpt}{B\textsubscript{K32}}
\newcommand{\hqarrf}{HQARRF-SC}

\title{Learning-Guided Planning in Large Dynamic Action Spaces:\\
Budgeted Tree Search for One-to-Many Mobile Charging}

\author{%
Liang-Ching Tao \and Pi-Chung Wang\thanks{Corresponding author.}\\
Department of Computer Science and Engineering, National Chung Hsing University\\
145 Xingda Rd., South District, Taichung 402, Taiwan\\
\texttt{liangkingtao@gmail.com}, \texttt{pcwang.tw@gmail.com}}

\date{Preprint, September 2026}

\begin{document}
\maketitle

\begin{abstract}
Many learned sequential decision systems map the current state directly to an
action.  That shortcut becomes brittle when candidate actions are numerous,
geometrically structured, and rebuilt with the state.  One-to-many mobile
charging makes this setting concrete: with $N{=}250$ sensors, the initial
state induces about $1{,}125$ candidate charging-stop actions; each chosen stop
simultaneously serves its in-range sensors, and the action universe changes as
sensors die.  LP-BTS is a \emph{learning-guided planning} architecture: a graph
proposal policy concentrates a small candidate support, a learned value critic
evaluates leaves, and edge-budgeted PUCT compares short simulated futures
before committing an action.  Because the policy scores this set without a
fixed output head, a single frozen checkpoint covers every evaluated setting,
spanning action universes from 736 to 2{,}813 stops.  Matched ablations reveal complementary effects:
uniform sampling costs 8.8 survival percentage points, while, with targeted
support fixed, PUCT jointly retains 1.4 points (about $3.5$ of $250$ sensors)
and direct policy selection travels 23\% farther.
On a prospectively specified, sealed 30-scenario confirmatory bank evaluated
once, LP-BTS attains the highest observed survival ($0.4545$) and alive-AUC
($0.8031$).  Its estimated survival advantage over the strongest
domain-engineered comparator is $+0.0066$ (95\% CI
$[-0.0037,\allowbreak +0.0184]$), an unresolved difference, while it exceeds a
deadline heuristic and two source-derived direct-policy reconstructions on
every paired scenario.  Both learned rows are trained, source-derived
reconstructions of variants reported by Gong et al.~\cite{gong2023}.  In this
setting, the results provide controlled evidence about learning-guided planning
in a large, dynamic action space.
\end{abstract}

\section{Introduction}\label{sec:intro}

Many sequential decision problems require choosing from action sets that are
large, structured, and state-dependent.  Direct state-to-action policies are
computationally attractive, but must encode downstream consequences in a
single mapping from the current state.  Explicit planning offers an
alternative: allocate limited computation to evaluate future trajectories
before committing an action.

One-to-many mobile charging makes the challenge tangible.  Wireless power
transfer lets a mobile charger (MC) replenish sensor nodes before they
die~\cite{kurs2007,xie2015,survey2022}; nodes drain continuously, the charger's
battery and travel time are finite, and deaths are irreversible.  Each action
selects a physical charging stop and simultaneously serves every sensor within
its radius.  For the central $N{=}250$-sensor system, geometry yields about
$1{,}125$ candidate stops at the initial decision.  The stop universe is then
rebuilt as sensors die, while every chosen stop commits hundreds of seconds of
travel, dwell, energy delivery, and continued sensor drain.

Brute-force planning is infeasible at this breadth, yet directly selecting one
stop asks a network to absorb all downstream consequences into a single score.
LP-BTS instead divides the work: a graph proposal policy identifies where to
spend search, a learned value critic evaluates shallow leaves, and
edge-budgeted tree search uses an explicit transition model to compare short
simulated futures.  The distinction is not whether long-term objectives are
learned, but whether plausible future trajectories are explicitly checked at
decision time.

This distinction is also motivated by a preregistered, method-independent
diagnostic on disjoint training worlds: under two fixed continuation policies,
the same candidate actions had near-zero rank correlation ($\rho=+0.0067$),
and a ridge diagnostic had held-out rank correlation $-0.137$.  These
measurements do not prove that a direct policy cannot learn; they show that,
under the diagnostic's short horizons and continuations, action quality was
strongly continuation-dependent.  Explicit lookahead is therefore a testable
response to that dependence rather than an assumed advantage
(\S\ref{sec:baselines}).

Learning-based WRSN schedulers have made real progress on this domain
problem~\cite{cao2021,gong2023,jiang2022,jiang2024,vuong2024,jnca2025}, but
their learned network generally commits the charging action directly.  In a
documented literature search (\S\ref{sec:related}) we found no prior WRSN
charging work that couples learned proposal and value models with explicit tree
search over charging actions.  This paper evaluates that architecture in
one-to-many mobile charging; one-to-many charging and dynamic stop locations
themselves have been studied in prior WRSN work~\cite{xie2015,gong2023}.  The
contributions are:

\begin{itemize}
\item \textbf{Dynamic action-space formulation.}  We formulate one-to-many
  mobile charging as planning over a state-dependent, geometrically structured
  charging-stop universe that can exceed $10^3$ actions while retaining the
  physical recipient set induced by each stop.  The policy carries no fixed
  output head, so one frozen checkpoint serves universes of 736 to 2{,}813
  stops across the evaluated grid, where a cell-based comparator holds its 471
  cells fixed and coarsens instead.
\item \textbf{Learning-guided budgeted planning.}  We combine a graph proposal
  policy, learned value critic, sampled candidate support, and edge-budgeted
  PUCT: learning concentrates computation on promising breadth, while
  explicit search checks short continuations before action commitment.
\item \textbf{Component gains and confirmatory evidence.}  Matched ablations
  quantify the distinct value of targeted proposal support and explicit search,
  characterize the budget/depth trade-off (\S\ref{sec:ablations}), and a
  prospectively specified sealed 30-scenario evaluation provides the principal
  result (\S\ref{sec:confirmatory}); all evaluation rules, method identities,
  contrasts, and confirmatory scenarios were fixed before that result existed.
\end{itemize}

We denote the final frozen configuration
\textbf{\ours{}} (Learned-Prior Budgeted Tree Search; internal checkpoint
identifier \ckpt{} in the released artifacts); its training procedure is
specified in \S\ref{sec:teacher}.  The sealed bank records the highest observed
survival/alive-AUC (0.4545/0.8031), though its difference from the strongest
domain-engineered comparator remains unresolved; development-grid evidence is
exploratory.

\section{Related Work}\label{sec:related}

\subsection{Heuristic and optimization-based WRSN charging}
Early work formulated charger routing as combinatorial optimization: Xie et
al.~\cite{xie2015} established multi-node (one-to-many) wireless charging with
cellular discretization and renewable-cycle optimization; a large literature
followed with on-demand architectures, spatial partitioning and priority
queues (see the survey~\cite{survey2022}).  Earliest-deadline ordering,
inherited from classical real-time scheduling~\cite{liu1973}, remains a strong
backbone: our K-EDF baseline (\S\ref{sec:baselines}) is a carefully engineered
$K$-node earliest-death-first scheduler in this family.  Crisis-aware
hybrid schedulers with per-episode tabular Q-learning --- our \hqarrf{} baseline~\cite{hqarrf}
--- represent the strongest handcrafted end of this spectrum in our
evaluation.

\subsection{Learning-based WRSN charging}
Cao et al.~\cite{cao2021} learn on-demand charging with time-window rewards.
Gong et al.~\cite{gong2023} --- the primary source for both of our learned
baselines --- cellularize the field into hexagons of the charging radius and
train a double-dueling DQN (OTM3DQN) to pick the next charging cell and
amount, explicitly targeting the one-to-many setting.  Attention-based
multi-agent actor--critic scheduling~\cite{jiang2022}, hybrid
discrete--continuous action spaces~\cite{jiang2024}, DQN-inspired adaptive
multi-node schemes~\cite{vuong2024}, and RL sequence scheduling for stochastic
event detection~\cite{jnca2025} extend this line.  The shared trait is direct
learned action selection: the network's forward pass \emph{is} the decision,
with any long-horizon information represented in its learned value or action
prediction.  Our method instead uses learned models inside explicit planning:
the proposal policy selects breadth, the critic evaluates leaves, and
budgeted search commits after simulated futures are compared.

\subsection{Learned planning and neural-guided search}
AlphaGo~Zero and AlphaZero~\cite{silver2017,silver2018} established a lineage
of policy-guided PUCT search~\cite{rosin2011} with self-generated training
targets; MuZero and Gumbel-based variants~\cite{schrittwieser2020,
danihelka2022} extended the recipe to learned models and sampled action
spaces.  They motivate the division of labor studied here, but not a claim of
cross-domain transfer: our $K{=}32$ proposals from $M\!\approx\!1{,}125$
candidates instantiate sampled-action-space search in a different, dynamic,
geometric domain.  We searched for prior WRSN applications
of Monte-Carlo tree search to charging scheduling and found none (the closest
use of MCTS in wireless power research optimizes circuit design, not
scheduling); we therefore describe the combination as not previously reported
for this problem in the literature we surveyed.

Table~\ref{tab:related} summarizes the axes that matter.

\begin{table*}[t]
\centering\small
\setlength{\tabcolsep}{4.5pt}
\caption{Related-work comparison.  ``Direct'' = the network's output is the
executed action; ``search'' = explicit lookahead commits the action.}
\label{tab:related}
\resizebox{\textwidth}{!}{%
\begin{tabular}{lllccll}
\toprule
Work & Problem & Action representation & 1-to-many & Learned & Decision & Relation to ours \\
\midrule
Xie et al.~\cite{xie2015} & lifetime optimization & cellular tour & yes & no & optimization & problem foundation \\
Cao et al.~\cite{cao2021} & on-demand charging & next node & no & yes & direct RL & learned baseline family \\
Gong et al.~\cite{gong2023} & online 1-to-many & hex cell + amount & yes & yes & direct RL (3DQN) & source of OTM3DQN/RMP baselines \\
Jiang et al.~\cite{jiang2022} & dynamic scheduling & node assignment & no & yes & direct MARL & attention/multi-agent line \\
Jiang et al.~\cite{jiang2024} & mobile charging & hybrid disc.+cont. & partial & yes & direct RL & richer actions, still direct \\
Vuong et al.~\cite{vuong2024} & large-scale charging & multi-node set & yes & yes & direct RL & DQN-inspired adaptive scheme \\
AlphaZero~\cite{silver2018} & games & moves & --- & yes & \textbf{search} & search recipe origin \\
Gumbel MuZero~\cite{danihelka2022} & games & sampled moves & --- & yes & \textbf{search} & sampled-action search \\
\textbf{Ours} & 1-to-many charging & stop point (+ recipient set) & yes & yes & \textbf{search} & only charging row with learned models \emph{inside} search \\
\bottomrule
\end{tabular}}
\end{table*}

\section{System Model}\label{sec:model}

$N$ sensors are deployed uniformly at random on a $W\times H$ field
($1000\times1000$\,m; $N{=}250$ centrally, $200$--$400$ in the sensitivity
grid).  Sensor $i$ holds energy $e_i(t)\in[0,E_{\max}]$ with $E_{\max}=150$,
drains at a base rate $\rho$ ($0.00375$\,units/s centrally) plus periodic
sensing and reporting costs, and dies irreversibly when $e_i(t)=0$.  In the
evaluated realizations every sensor communicates \emph{directly} with the base
station at the field center (the stored topology is a star; measured relay
depth is exactly one), so no multi-hop relay-load effects are present or
claimed in this study.

A single MC with battery capacity $C$ ($10{,}000$ centrally) travels at speed
$v$ ($5$\,m/s), consumes energy both moving and charging, and recharges itself
at the base station at power $50$.  Charging is \emph{one-to-many}: when the
MC dwells at stop point $p$, every alive sensor within radius $R$ ($30$\,m)
receives power $P_c$ ($10$ centrally) at efficiency $\eta=0.9$
simultaneously.  Decisions are event-driven: whenever the MC completes its
current activity, the scheduler chooses its next charging stop.

\textbf{Return-to-base semantics.}  Returning to base is \emph{not} a
scheduler action.  The environment enforces an energy-reserve rule: if the
MC's residual energy cannot cover the commanded movement \emph{plus} the cost
of returning to base afterwards, the environment overrides the command and
sends the MC home to recharge.  All schedulers in this paper --- learned,
heuristic, and ours --- operate under this same rule.

\textbf{Metrics.}  Primary: \emph{survival} $S(T)$, the fraction of sensors
alive at horizon $T{=}30{,}000$\,s.  Secondary: time-normalized alive-AUC,
total MC travel distance, and per-decision computation time.

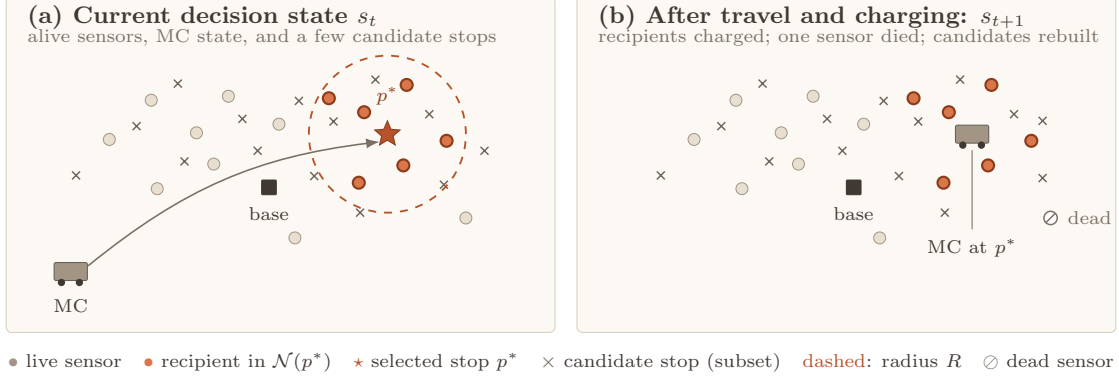
\begin{figure*}[t]
\centering
\begin{tikzpicture}[
  x=0.88cm, y=0.88cm,
  every node/.style={font=\scriptsize, text=lpInk},
  sensor/.style={circle, draw=lpSandEdge, line width=.35pt, minimum size=4.5pt,
                 inner sep=0pt},
  live/.style={sensor, fill=lpSand},
  recipient/.style={sensor, fill=lpAccentFill, draw=lpAccentDark, line width=.8pt},
  candidate/.style={lpCross, line width=.55pt},
  mc/.style={fill=lpSandEdge, draw=lpInk!80, rounded corners=.7pt},
  arr/.style={-{Latex[length=1.8mm]}, line width=.60pt, lpMuted}]

\draw[rounded corners=2pt, fill=lpPaper, draw=lpLine] (0,0) rectangle (8.25,5.05);
\node[anchor=west, font=\small\bfseries] at (0.18,4.74) {(a) Current decision state $s_t$};
\node[anchor=west, text=lpMuted] at (0.18,4.43) {alive sensors, MC state, and a few candidate stops};

\draw[fill=lpInk, draw=lpInk, rounded corners=.4pt] (3.84,2.07) rectangle +(0.22,0.22);
\node[below=1pt] at (3.95,2.07) {base};
\draw[mc] (0.72,0.78) rectangle +(0.50,0.27);
\fill[lpInk] (0.83,0.75) circle (.055) (1.10,0.75) circle (.055);
\node[below=2pt, align=center] at (0.97,0.72) {MC};

\foreach \x/\y in {1.55/2.90,2.18/3.49,2.86/3.00,3.34/3.55,4.10/3.13,2.27/2.16,3.12/2.53,4.34/1.42,6.91/1.72}{
  \node[live] at (\x,\y) {};
}

\foreach \x/\y in {1.05/2.36,1.96/3.10,2.58/3.74,2.68/2.57,3.55/3.21,3.78/2.73,4.40/3.48,4.92/3.17,5.55/3.80,5.73/2.98,6.36/3.28,6.58/2.22,7.19/2.73,4.63/2.35,5.32/1.80}{
  \draw[candidate] (\x-.06,\y-.06)--(\x+.06,\y+.06) (\x-.06,\y+.06)--(\x+.06,\y-.06);
}

\draw[dashed, line width=.7pt, lpAccent] (5.73,2.98) circle (1.18);
\foreach \x/\y in {4.85/3.52,5.38/3.31,6.02/3.72,6.62/2.88,5.30/2.25,5.97/2.51}{
  \node[recipient] at (\x,\y) {};
}
\node[star, star points=5, star point ratio=2.2, draw=lpAccentDark,
      fill=lpAccent, minimum size=10pt, inner sep=0pt] (chosen) at (5.73,2.98) {};
\draw[arr, bend left=16] (1.23,1.00) to (chosen.south west);
\node[above=2pt, font=\scriptsize\bfseries, text=lpAccentDark] at (chosen.north) {$p^\ast$};

\draw[rounded corners=2pt, fill=lpPaper, draw=lpLine] (8.58,0) rectangle (16.68,5.05);
\node[anchor=west, font=\small\bfseries] at (8.76,4.74) {(b) After travel and charging: $s_{t+1}$};
\node[anchor=west, text=lpMuted] at (8.76,4.43) {recipients charged; one sensor died; candidates rebuilt};
\draw[fill=lpInk, draw=lpInk, rounded corners=.4pt] (12.63,2.07) rectangle +(0.22,0.22);
\node[below=1pt] at (12.74,2.07) {base};

\foreach \x/\y in {10.34/2.90,10.97/3.49,11.65/3.00,12.13/3.55,12.89/3.13,11.06/2.16,11.91/2.53,13.13/1.42}{
  \node[live] at (\x,\y) {};
}
\foreach \x/\y in {9.84/2.36,10.75/3.10,11.37/3.74,11.47/2.57,12.34/3.21,12.57/2.73,13.19/3.48,13.71/3.17,14.34/3.80,15.15/3.28,13.42/2.35,14.11/1.80,15.58/3.18,15.58/2.32}{
  \draw[candidate] (\x-.06,\y-.06)--(\x+.06,\y+.06) (\x-.06,\y+.06)--(\x+.06,\y-.06);
}
\foreach \x/\y in {13.64/3.52,14.17/3.31,14.81/3.72,15.41/2.88,14.09/2.25,14.76/2.51}{
  \node[recipient] at (\x,\y) {};
}
\draw[lpMuted, fill=white, line width=.55pt] (15.70,1.72) circle (.10);
\draw[lpMuted, line width=.55pt] (15.63,1.65)--(15.77,1.79);
\node[anchor=west, text=lpMuted] at (15.82,1.72) {dead};

\draw[mc] (14.27,2.845) rectangle +(0.50,0.27);
\fill[lpInk] (14.38,2.815) circle (.055) (14.66,2.815) circle (.055);
\draw[lpMuted, line width=.4pt] (14.52,1.55) -- (14.52,2.74);
\node[anchor=north] at (14.52,1.55) {MC at $p^\ast$};

\node[anchor=north, font=\scriptsize] at (8.34,-0.17)
  {\textcolor{lpSandEdge}{$\bullet$}~live sensor\quad
   \textcolor{lpAccentFill}{$\bullet$}~recipient in $\mathcal N(p^\ast)$\quad
   \textcolor{lpAccent}{$\star$}~selected stop $p^\ast$\quad
   \textcolor{lpCross}{$\times$}~candidate stop (subset)\quad
   \textcolor{lpAccent}{dashed}: radius $R$\quad
   \textcolor{lpMuted}{$\oslash$}~dead sensor};
\end{tikzpicture}
\caption{Schematic LP-BTS decision state (illustrative; not an evaluated
scenario and not to scale).  (a)~An action is a physical charging stop
$p\in\mathcal{A}(s_t)$; crosses show a few stops that
Algorithm~\ref{alg:cands} induces from the alive geometry
($M\approx1{,}125$ at $N{=}250$).  Executing the selected stop $p^\ast$
charges every live sensor within radius $R$ (dashed circle), i.e., its
recipient set $\mathcal N(p^\ast)$ (filled, thick-outlined).  (b)~After travel,
charging, and drain, one sensor has died; the stop universe is rebuilt from
the alive set at every decision.  Stops are neither cluster centres nor fixed
grid cells; Figure~\ref{fig:pipeline} shows how $p^\ast$ is selected.}
\label{fig:decision-map}
\end{figure*}

\section{Problem Formulation}\label{sec:problem}

At decision epoch $t$ the state $s_t$ comprises all sensor positions,
energies and alive flags, the MC position and residual energy, and time.  The
action space $\mathcal{A}(s_t)$ is the set of \emph{charging stops}: candidate
points $p$ derived from the geometry of the alive set
(\S\ref{sec:cands}), each inducing the recipient set
$\mathcal{N}(p)=\{i: \lVert x_i-p\rVert\le R,\; e_i>0\}$.  Executing
$a=p$ advances the simulator through travel and dwell physics --- including,
when the reserve rule fires, a forced base return.  The objective is
$\max\,\mathbb{E}[S(T)]$.

Figure~\ref{fig:decision-map} illustrates this action representation with a
schematic state rather than an evaluation snapshot: a stop is a physical point
that serves its entire in-range recipient set, and the stop universe is rebuilt
from the alive set at every decision.

This is a large, dynamic, structured, long-consequence action space.  It is
\emph{large}: $|\mathcal{A}(s_t)|$ is about $1{,}125$ initially at $N{=}250$
and exceeds $2{,}800$ at $N{=}400$.  It is \emph{dynamic}: the universe is
reconstructed from the alive set and changes as sensors die.  It is
\emph{structured}: actions are geometric stops inducing recipient sets, not
independent categorical labels.  It has \emph{long consequences}: executing a
stop commits travel, dwell time, energy delivery, sensor drain, and possibly a
future forced return to base.  A fixed output head over a static
discretization either coarsens or freezes this space --- measurably so: tying
the discretization to the charging radius swings a comparator's output width
from 471 to 4{,}012 actions, and its parameter count by $8.5\times$, before any
scheduling question is asked (\S\ref{sec:baselines}).  Explicit lookahead can
instead evaluate the resulting short trajectories at decision time.

\section{Proposed Method}\label{sec:method}

\begin{figure*}[t]
\centering
\resizebox{0.98\textwidth}{!}{%
\begin{tikzpicture}[
  font=\small,
  node distance=4mm and 6mm,
  box/.style={draw, rounded corners=2pt, align=center, inner sep=5pt, minimum height=10mm},
  data/.style={box, fill=lpNeutral, draw=lpInk},
  learned/.style={box, fill=lpSage, draw=lpSageEdge},
  frozen/.style={box, fill=lpSage, draw=lpSageEdge},
  proc/.style={box, fill=lpNeutral, draw=lpInk},
  arr/.style={-{Latex[length=2.6mm]}, thick, lpInk}]
\node[data] (state) {decision state $s_t$\\ \scriptsize positions, energies,\\ \scriptsize MC battery, time};
\node[proc, right=of state] (cand) {large dynamic action universe\\ \scriptsize Algorithm~\ref{alg:cands}\\ \scriptsize initial $M\approx 1{,}125$ stops};
\node[learned, right=of cand] (pol) {proposal policy $\pi_\theta$\\ \scriptsize learning-guided support};
\node[proc, right=of pol] (sample) {sample $K{=}32$ from $\beta$\\ \scriptsize multiplicities $\to$ corrected prior\\ \scriptsize planning support};
\node[proc, right=of sample, fill=lpClay, draw=lpAccentDark] (search) {budgeted planning: PUCT\\ \scriptsize simulated futures, $B_{\mathrm{edge}}{=}2{,}048$\\ \scriptsize transition model + base returns};
\node[data, right=of search, fill=lpClay, draw=lpAccentDark, line width=.8pt] (act) {committed stop $p^\ast$\\ \scriptsize recipients $\mathcal{N}(p^\ast)$};
\node[frozen, below=5mm of search] (critic) {frozen learned critic $V_\phi$\\ \scriptsize leaf evaluation};
\draw[arr] (state) -- (cand);
\draw[arr] (cand) -- (pol);
\draw[arr] (pol) -- (sample);
\draw[arr] (sample) -- (search);
\draw[arr] (search) -- (act);
\draw[arr] (critic) -- (search);
\end{tikzpicture}}
\caption{One decision of \ours{} as learning-guided planning.  The proposal
policy identifies where to search, the learned critic evaluates leaves, and
budgeted search over simulated futures commits the action.}
\label{fig:pipeline}
\end{figure*}
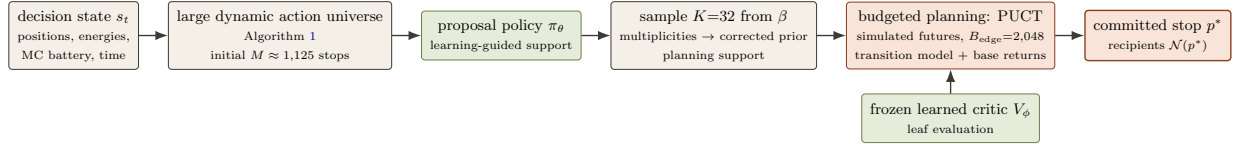

Figure~\ref{fig:pipeline} shows the learning-guided planning system.  \ours{}
constructs a structured candidate universe (\S\ref{sec:cands}); the graph
proposal policy $\pi_\theta$ determines \emph{where} to search, sampling forms
the support, the transition model determines \emph{how} futures evolve, the
value critic $V_\phi$ evaluates leaves, and budgeted PUCT allocates
computation before committing an action.  Thus both breadth selection and
terminal evaluation are learned, while tree search supplies explicit
lookahead.  Sampling, search, and the frozen critic are specified in
\S\ref{sec:sampling}--\ref{sec:teacher}.

\subsection{Canonical one-to-many stop universe}\label{sec:cands}

The action space is rebuilt at every decision from the alive set
$\mathcal{V}=\{i: e_i>0\}$ (Algorithm~\ref{alg:cands}).  Geometric proposals
come from four primitive families over a $k$-nearest-neighbour graph
($k{=}8$): each alive sensor's own position (\textsf{atomic}); for each
neighbouring pair within distance $2R$, the pair midpoint
(\textsf{midpoint}) and the two intersection points of the radius-$R$ circles
centred on the pair (\textsf{intersection}); and for each neighbouring triple
whose minimum enclosing circle has radius $\le R$, that circle's centre
(\textsf{triple\_center}).  Proposals are then \emph{canonicalized}:
coordinates are quantized to a fixed grid quantum and all proposals sharing a
quantized coordinate collapse into one action whose identity is the physical
stop point alone.  The generating pair/triple (the ``planned members'') is
retained only as diagnostic provenance --- it never distinguishes actions,
so geometrically equivalent groups cannot create non-causal duplicate
actions: the physics of dwelling at $p$ depends only on $p$, and the
recipient set $\mathcal{N}(p)$ is recomputed from the distance matrix, not
from the proposal's source.  No coverage-set collapse, priority filtering, or
cap is applied: the search sees the complete implemented universe --- every
canonical stop induced by the kNN construction --- ordered by canonical stop
key.  Pairs and triples are drawn from kNN neighbourhoods ($k{=}8$), so this
is the full enumeration of the defined construction, not of every geometric
stop in the plane.  This $k{=}8$ graph serves candidate geometry only; it is
intentionally distinct from the $k{=}12$ graph used by the neural encoder for
message passing.

Complexity is $O(|\mathcal{V}|\,k)$ pairs and $O(|\mathcal{V}|\,k^2)$ triples
for generation plus an $M\times|\mathcal{V}|$ distance matrix for recipient
closure.  Measured at the first decision of the stored scenarios, $M{=}736$ at
$N{=}200$, $1{,}125$ at $N{=}250$, and $2{,}813$ at $N{=}400$; the universe
grows with the charging radius (306 at $R{=}10$\,m to $1{,}560$ at
$R{=}35$\,m) and shrinks with the alive set as sensors die.

\begin{algorithm}[t]
\caption{Canonical charging-stop universe (as implemented)}
\label{alg:cands}
\begin{algorithmic}[1]
\raggedright
\Require alive set $\mathcal{V}$, positions $x$, radius $R$, kNN degree $k{=}8$
\State $P \gets \{(\textsf{atomic}, x_i) : i \in \mathcal{V}\}$
\ForAll{kNN pairs $(i,j)$, $\lVert x_i-x_j\rVert \le 2R$}
  \State $P \gets P \cup \{(\textsf{midpoint}, \tfrac{x_i+x_j}{2})\}$
  \Statex \hspace{\algorithmicindent}$\cup\ \{(\textsf{intersection}, q^\pm_{ij})\}$
          \Comment{circle intersections}
\EndFor
\ForAll{kNN triples $\tau$ with $\mathrm{MEC}(\tau).r \le R$}
  \State $P \gets P \cup \{(\textsf{triple\_center}, \mathrm{MEC}(\tau).c)\}$
\EndFor
\State group $P$ by quantized coordinates; keep one representative per group
\State \Return representatives sorted by canonical stop key; for each stop
       $p$, recipients $\mathcal{N}(p)=\{i\in\mathcal{V}:\lVert x_i-p\rVert\le R\}$
\end{algorithmic}
\end{algorithm}

\subsection{Graph policy and value networks}\label{sec:nets}

Both networks are small graph neural networks (GNNs) built on the same graph
description of the state; the $k$-nearest-neighbour (kNN) construction supplies
graph neighbourhoods rather than serving as a kNN classifier or regressor.
Sensors are
nodes with nine features: normalized $x$ and $y$ coordinates, energy fraction,
capacity, consumption rate, clipped time-to-death, alive flag, distance to the
charger, and normalized current array index.  A $k$-nearest-neighbour graph
($k{=}12$) carries five edge features: relative $x$ and $y$ offset, distance,
an in-charging-range flag, and the consumption-rate difference.  Six global
features summarize time, charger $x$ and $y$ position, charger energy, mean
alive energy fraction, and the alive fraction.  A candidate stop is described
by six features: its $x$ and $y$ position, distance from the charger, in-range
recipient count, a has-recipient flag, and the in-range energy deficit it
would cover; it also has a per-sensor relation tensor (relative $x$ and $y$
offset, in-range flag, distance, deficit).  Thus the policy scores a
\emph{dynamic} set of $M$ stops without a fixed output head.%
\footnote{The sensor feature vector includes the current array index, so we
claim no equivariance to arbitrary sensor relabeling.}

\textbf{Encoder.}  Both networks use the same encoder architecture with
separate weights.  Linear layers project sensor, edge and global features to a
width of 64.  One round of edge-conditioned message passing runs over a
directed $k$-nearest-neighbour graph ($k{=}12$, no self-loops): each sensor
sums $\tanh(W_s x_i + W_e e_{ij})$ over its in-neighbours and adds
$\tanh(W_l\,\cdot)$ of the scaled sum back to its own embedding (a residual
update).  One multi-head self-attention layer (4 heads) over all sensors, also
residual, follows.  A graph vector $g$ is the mean sensor embedding plus a
linear projection of the six global features.  There is no normalization
layer and no dropout.

\textbf{Candidate scoring (policy).}  For every candidate stop $a$ the
network forms a relation embedding $r_{aj} = s_j + W_r\,\phi(a,j)$ over all
sensors $j$, where $\phi$ is the five-dimensional candidate--sensor relation.
Two max-pools over sensors summarize it --- one unmasked, one restricted to
the alive sensors within charging range of $a$ (a learned null vector when
$a$ covers none) --- and are concatenated with a linear embedding of the six
candidate features and with $g$.  A two-layer MLP ($256\!\to\!64\!\to\!1$,
ReLU) yields one logit per candidate; candidates are processed in chunks of
256 with shared weights, so reordering candidates reorders the corresponding
scores and the scorer handles any $M$.  One checkpoint therefore served every
setting reported here, from $M{=}736$ at $N{=}200$ to $M{=}2{,}813$ at
$N{=}400$, with no per-geometry retraining.  $\pi_\theta(\cdot\mid s)$ is the
softmax over the $M$ logits.

\textbf{Value (critic).}  $V_\phi(s)=\sigma(\mathrm{MLP}([\bar s;\,g]))$ with
a $128\!\to\!64\!\to\!1$ ReLU MLP over the mean sensor embedding and the graph
vector; the sigmoid bounds the output to $(0,1)$, the scale of terminal
survival.  The policy has 39{,}681 parameters and the critic 30{,}593
(Table~\ref{tab:arch}); both checkpoints match these modules tensor for
tensor.

\begin{table}[t]
\centering\small\setlength{\tabcolsep}{5pt}
\caption{Network specification (frozen implementation).}
\label{tab:arch}
\footnotesize
\begin{tabular}{@{}l P{0.60\columnwidth}@{}}
\toprule
input features & sensor 9, edge 5, global 6, candidate 6, candidate--sensor relation 5 \\
graph & directed kNN, $k{=}12$, no self-loops, rebuilt per state \\
width & 64 \\
message passing & 1 layer, edge-conditioned sum, $\tanh$, residual \\
attention & 1 layer, 4 heads, full self-attention, residual \\
norm.\ / dropout & none / none \\
policy head & two max-pools (global, masked-local) $\to$ MLP $256\!\to\!64\!\to\!1$, ReLU \\
critic head & MLP $128\!\to\!64\!\to\!1$, ReLU, sigmoid output \\
parameters & policy 39{,}681; critic 30{,}593 \\
\bottomrule
\end{tabular}
\end{table}

\subsection{Proposal sampling and sampling-corrected prior}\label{sec:sampling}

The policy produces logits over all $M$ candidates;
$\pi = \mathrm{softmax}(\mathrm{logits})$.  Proposals are $K$ i.i.d.\ draws
with replacement from the tempered, smoothed distribution
\begin{equation}\label{eq:beta}
\beta_a \;=\; (1-\epsilon)\,
\frac{\pi_a^{1/\tau}}{\sum_{b=1}^{M}\pi_b^{1/\tau}}
\;+\; \frac{\epsilon}{M},
\qquad \tau{=}1,\ \epsilon{=}0.05,
\end{equation}
collapsing to $\le K$ unique arms (typically 24--32 of $K{=}32$).  At the
initial decision of the central setting these $K{=}32$ draws sample at most
${\sim}2.8\%$ of the $M{\approx}1{,}125$ candidates, so \emph{proposal quality} --- not the size
of the universe --- governs what the search ever evaluates, which is why the
learned prior dominates the ablations (\S\ref{sec:ablations}).  With
empirical draw frequencies $\hat\beta_a = c_a/K$ on the sampled support, the
search prior is
\begin{equation}\label{eq:peff}
P_{\mathrm{eff}}(a) \;=\;
\frac{(\hat\beta_a/\beta_a)\,\pi_a}
     {\sum_{b\in\mathrm{support}} (\hat\beta_b/\beta_b)\,\pi_b}.
\end{equation}
Dividing the empirical draw frequency $\hat\beta_a$ by the sampling
probability $\beta_a$ of Eq.~\eqref{eq:beta} compensates for the proposal distribution induced by
tempering and $\epsilon$-smoothing, so the prior the search uses tracks the policy
$\pi$ rather than the sampling distribution $\beta$ (for $\tau{=}1,\epsilon{=}0$
it reduces to the empirical draw frequencies).  We call it a
\emph{sampling-corrected} prior: as $K$ grows, $\hat\beta$ approaches $\beta$
and the corrected prior approaches the policy weighting on the sampled
support.%
\footnote{The normalization in Eq.~\eqref{eq:peff} is self-normalized, so we
make no finite-sample unbiasedness claim.}

\subsection{Edge-budget search}\label{sec:search}
PUCT ($c_{\mathrm{puct}}{=}1.5$) with the sampling-corrected prior
$P_{\mathrm{eff}}$ of Eq.~\eqref{eq:peff} expands a tree over the sampled arms, where
every tree-edge traversal executes one simulated environment transition
(travel, dwell, drain --- and forced base returns, which the internal model
mirrors exactly from the environment rule).  The budget is
$B_{\mathrm{edge}}{=}2{,}048$ transitions per decision with a safety depth
ceiling of 16 that is never binding (measured mean leaf depth $2.19$, maximum
5): under this budget the search chooses breadth.  Leaves are evaluated by
the frozen critic.  The visit-count argmax is executed.  A per-decision seeded
RNG ($\mathrm{seed}=300+\text{decision index}$) makes evaluation
bit-reproducible within a fixed platform/runtime stack; we verified that
changing the evaluation seed does not change any outcome on that stack.

\subsection{Training procedure}\label{sec:teacher}

\begin{figure}[t]
\centering
\resizebox{\columnwidth}{!}{%
\begin{tikzpicture}[
  font=\small, node distance=4.5mm,
  stage/.style={draw, rounded corners=2pt, align=center, inner sep=4pt,
                minimum height=8mm, minimum width=52mm},
  arr/.style={-{Latex[length=2.4mm]}, thick, lpInk}]
\node[stage, fill=lpNeutral, draw=lpInk] (expert) {search-based expert (teacher)};
\node[stage, fill=lpSage, draw=lpSageEdge, below=of expert] (A) {bootstrap policy checkpoint $A$};
\node[stage, fill=lpNeutral, below=of A, draw=lpInk] (cut)
  {\textbf{permanent teacher cutoff}};
\node[stage, fill=lpClay, draw=lpAccentDark, below=11mm of cut] (B)
  {\textbf{\ours{}} --- final frozen configuration (\ckpt)};
\draw[arr] (expert) -- node[right]{imitation} (A);
\draw[arr] (A) -- (cut);
\draw[arr] (cut) -- node[right, align=left, font=\scriptsize, xshift=1mm]{one autonomous\\ teacher-free update} (B);
\node[draw=lpSageEdge, fill=lpSage, rounded corners=2pt, align=center,
      inner sep=3pt, font=\scriptsize, below=4mm of B, minimum width=52mm] (critic)
  {critic $V_\phi$: trained in an earlier generation, then frozen;\\
   byte-identical across $A\!\to\!$\ours{} and all evaluation};
\draw[arr, dashed] (critic.west) to[out=180,in=180] (A.west);
\draw[arr, dashed] (critic.east) to[out=0,in=0] (B.east);
\end{tikzpicture}}
\caption{Training pipeline: expert bootstrap, permanent teacher cutoff, one
teacher-free update.  The $A\!\to\!$\ours{} step is a training property, not
a measured improvement (see text).}
\label{fig:training}
\end{figure}
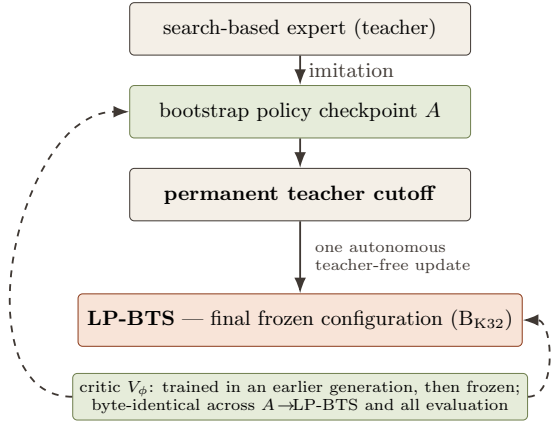

The proposal policy is bootstrapped by imitation of a search-based expert
(Fig.~\ref{fig:training}).  The teacher is then removed permanently: after checkpoint $A$, no expert
labels, expert queries, or teacher-derived signals of any kind exist anywhere
in training or evaluation.  The method then performs one complete autonomous
policy update --- self-generated search visit distributions as targets, the
critic held frozen --- yielding \ours{}.  At evaluation time the serving stack
consists solely of the frozen policy and critic checkpoints inside the search;
there is no teacher in the loop.

\textbf{Expert and bootstrap.}  The teacher is a handcrafted stop-point MCTS
planner (64 simulations of depth 3 per decision, PUCT with a hand-designed
leaf evaluator and urgency prior, 32-candidate cap).  It played 120 episodes
on the central-physics scenario family (seeds 1000--1119, disjoint from every
evaluation bank).  For each of its decisions we recorded the executed stop
and its 32-action candidate set, and trained a fresh policy with a listwise
target over that set plus 224 uniformly drawn negatives from the full
universe (executed stop $0.5$, remaining teacher candidates share $0.5$,
negatives $0$) under a sampled-softmax cross-entropy: Adam, learning rate
$10^{-3}$, 8 epochs, one decision per step, an episode-level split holding
out every fifth seed (8{,}222 training decisions), final-epoch weights kept.
This is checkpoint $A$.

\textbf{Cutoff and teacher-free update.}  After $A$ the expert is never
queried again.  $A$ (with the frozen critic) played 40 self-play episodes
(seeds 800--839; $B_{\mathrm{edge}}{=}2{,}048$ chosen beforehand by a fixed
selection rule between 1{,}024 and 2{,}048 on five separate seeds), yielding
2{,}861 decisions.  The policy was then warm-started from $A$ and fitted to
the search's own root visit distribution over the $\le 32$ sampled arms
(again with 224 uniform negatives) under the same sampled-softmax
cross-entropy, Adam $10^{-3}$, 8 epochs, 2{,}288 training decisions and a
fixed 200-decision validation set; the final-epoch weights are the \ours{}
policy.  Validation cross-entropy to the search targets fell from 3.51
(checkpoint $A$) to 3.21.

\textbf{Critic.}  $V_\phi$ was trained in an earlier development generation
from 79 self-play episodes (about 3{,}300 states, at most 60 per episode)
under a different search operator, with the episode's terminal survival as
the regression target (MSE, Adam $10^{-3}$, batch 32, gradient clipping 5,
8 epochs, best-validation checkpoint; held-out MAE 0.026).  A refit of the
critic on the method's own episodes provides a closed-loop control: although
it improved offline ranking metrics, it left survival unchanged ($-0.0008$,
95\% CI $[-0.0053,\allowbreak +0.0040]$, $n{=}30$ development scenarios) and
increased travel.  The pre-specified rule therefore retained the simpler,
frozen critic.  Table~\ref{tab:training} summarizes; the released
artifacts record every checkpoint identity.

\begin{table}[t]
\centering\footnotesize\setlength{\tabcolsep}{4pt}
\caption{Training procedure (frozen; all on central physics).}
\label{tab:training}
\begin{tabular}{@{}l P{0.25\columnwidth} P{0.25\columnwidth} P{0.23\columnwidth}@{}}
\toprule
 & bootstrap ($A$) & teacher-free update (\ours{}) & critic $V_\phi$ \\
\midrule
data & 120 expert episodes, 8{,}222 dec. & 40 self-play episodes, 2{,}288 dec. & 79 episodes, ${\sim}$3{,}300 states \\
target & listwise: executed 0.5 + teacher set 0.5 & search visit distribution & terminal survival \\
loss & sampled-softmax CE (+224 neg.) & sampled-softmax CE (+224 neg.) & MSE \\
optimizer & Adam $10^{-3}$, 8 ep., 1 dec./step & Adam $10^{-3}$, 8 ep., 1 dec./step & Adam $10^{-3}$, 8 ep., batch 32 \\
init & fresh & warm start from $A$ & fresh \\
selection & final epoch & final epoch & best validation \\
\bottomrule
\end{tabular}
\end{table}

\ours{} was fixed as the final configuration before the sensitivity grid was
evaluated; $A$ is reported as an ablation.  On the central development
scenarios, paired per scenario, $S(\text{\ours{}})-S(A) = -0.0012$ (95\% CI
$[-0.0108,\allowbreak +0.0096]$, \rpun{} $3/2/5$,
$n{=}10$): the teacher-free update yields a final policy whose inference is
fully independent of the expert, without a measurable change in closed-loop
survival.

\section{Experimental Methodology}\label{sec:methodology}

\subsection{Frozen paired benchmark}
All development-grid evaluation uses a frozen bank of 300 stored scenario realizations
(``worlds''): 30 families (one per factor level, 6 axes) $\times$ 10
scenarios, each scenario's
sensor coordinates, initial energies, topology and workload phases stored
literally (not regenerated from seeds).  Families share their 10 underlying
random realizations --- verified by masked byte-comparison, every family is
the identical draw with only the factor changed --- so every factor curve is a
\emph{paired} comparison on a common scenario spine.  Evaluation is
event-accurate simulation to $T{=}30{,}000$\,s and was empirically
deterministic under the measured runtime configuration.  The separately specified 30-scenario sealed confirmatory bank
is described in \S\ref{sec:confirmatory}.

In addition to scenario identity, the constructed physical state was hashed
before evaluation: all 250 LP-BTS development rows matched the corresponding
baseline fingerprint exactly, and every central ablation arm passed the same
check on all ten worlds.  Thus a method comparison changes policy logic, not
simulated physics.

\subsection{Baselines}\label{sec:baselines}
\textbf{\hqarrf{}}~\cite{hqarrf}: a strong crisis-aware hybrid scheduler with
per-episode tabular Q-learning from our prior work; the strongest baseline throughout.
Appendix~\ref{app:hqarrf} specifies the executed single-charger variant in full.
\textbf{K-EDF}: a $K$-node earliest-death-first scheduler in the classical
EDF family~\cite{liu1973}, engineered by us as a strong deterministic
baseline.  \textbf{OTM3DQN}~\cite{gong2023}: documented source-derived
reconstruction of the one-to-many double-dueling DQN, adapted only where
required by the common physics and passed through a frozen 16-item conformance
contract.  Its final checkpoints follow the paper-derived frozen protocol
(3 seeds $\times$ 1{,}200 charging rounds).
\textbf{RMP-RL-cell}~\cite{gong2023,cao2021}: the cell-based RL comparator
applied in~\cite{gong2023}, reconstructed under the same common-physics
adaptations and conformance contract.  Its final checkpoints likewise follow
the frozen paper-derived protocol (3 seeds $\times$ 1{,}200 charging rounds).
\textbf{NULL}: the MC never moves (floor).

Both reconstructed learned rows are trained-policy evaluations of variants
reported in the same source study~\cite{gong2023}: each of three frozen seeds
followed the paper-derived 1{,}200-round protocol in the common simulator, and
the resulting frozen checkpoints are compared with \ours{} on matched
scenarios (Table~\ref{tab:confirmatory}).  A charging round ends when the MC
completes its base-charge cycle; it is not a single decision.  Training and
validation worlds are disjoint, and the sealed bank is excluded from training.

\textbf{Why the grid fixes the charging radius.}  Both reconstructed
comparators cellularize the field into cells of the charging radius, so the
radius fixes the lattice, the lattice fixes the action set, and the action set
fixes the network's output width: the OTM3DQN architecture instantiates 471
cells and $2.1$M parameters at $R{=}30$\,m but 4{,}012 cells and $17.6$M
parameters at $R{=}10$\,m.  A checkpoint trained at one radius therefore cannot
be loaded at another, and an in-distribution radius panel would need a
separately trained family per radius value.  \ours{} carries no such
constraint: its policy has no fixed output head and rebuilds the stop universe
from the live geometry at every decision (\S\ref{sec:cands}), so one frozen
checkpoint applies unchanged as the universe grows from 306 stops at
$R{=}10$\,m to $1{,}560$ at $R{=}35$\,m.  We nevertheless report no \ours{}
rows on that axis --- with no learned comparator able to appear beside them the
panel was not run --- and the five perturbed-radius families are held out of
the 25-setting grid throughout.  Training one comparator family per radius
would not repair the comparison: the resulting networks are different models,
ranging over an $8.5\times$ span in capacity, so a radius curve built that way
would vary the comparator's identity along its own $x$-axis.  Holding every
method's identity frozen across the grid is what makes the remaining 25
settings a paired comparison at all.

\textbf{A refuted prior, retained.}  A preregistered baseline sanity audit
expected a positive-or-saturating charging-radius trend.  Its measured trend
was instead negative monotone and is reported as refuted rather than relabelled
post hoc; travel rose with radius, a coherent but not conclusive mechanism.
We make no LP-BTS radius-performance claim.

\subsection{Statistical procedure}\label{sec:stats}
All rules were fixed before any result for \ours{} existed.  Differences are
computed per paired scenario, never as differences of group means.  For
deterministic methods the scenario realization is the statistical unit; for
\hqarrf{}, whose behaviour depends on an algorithm seed, the three seeds are
averaged within each scenario first, so its unit is also the scenario.  For
the two learned baselines the training seed is the primary unit ($n{=}3$):
paired differences are formed per (training seed, scenario), averaged within
the seed, and we report the wider of a Student-$t$ interval and a seed-level
cluster-bootstrap interval.  World-unit intervals are 95\% paired cluster
bootstraps (10{,}000 resamples, fixed seed); a one-sensor materiality floor
($1/N_0{=}0.004$) is applied in the signed direction.  Every row within a
table comes from one engine on one platform.  Development setting-level
verdicts are exploratory: 25 individual intervals are reported without
multiplicity correction, and the full per-setting table
(Table~\ref{tab:grid}) lets readers apply any correction they prefer.
Appendix~\ref{app:stats} gives the complete specification.

\section{Development and Sensitivity Results}\label{sec:results}

\begin{table}[t]
\centering\small
\caption{Verdicts across the 25 evaluated non-radius settings (exploratory).
Classical and \hqarrf{} rows use scenario-level paired intervals; reconstructed
learned-comparator rows use the frozen training-seed hierarchy ($n{=}3$).
Resolved positive/negative
means the corresponding 95\% CI excludes zero after applying the signed
one-sensor materiality rule of \S\ref{sec:stats}.}
\label{tab:tally}
\resizebox{\columnwidth}{!}{%
\begin{tabular}{lccc}
\toprule
\ours{} vs. & Resolved $+$ & Unresolved & Resolved $-$ \\
\midrule
K-EDF & \textbf{25} & 0 & 0 \\
OTM3DQN & \textbf{25} & 0 & 0 \\
RMP-RL-cell & \textbf{25} & 0 & 0 \\
NULL & \textbf{25} & 0 & 0 \\
\hqarrf{} & 4 & 18 & 3 \\
\bottomrule
\end{tabular}}
\end{table}

\begin{figure*}[t]
\centering
\includegraphics[width=\textwidth]{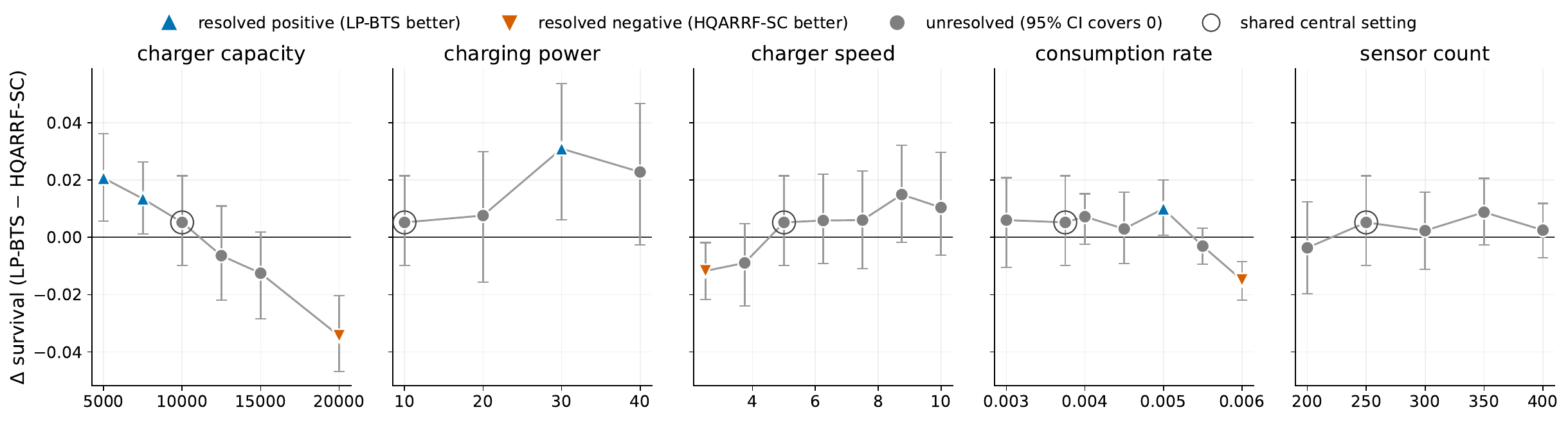}
\caption{Regime sensitivity (development grid, exploratory): paired
survival difference \ours{} $-$ \hqarrf{} along five factor axes (10
scenarios per point, 95\% paired bootstrap CIs).  Marker shape encodes the
verdict (up = resolved positive, down = resolved negative, circle =
unresolved); the ring marks the shared central setting.  The development-grid
means vary by axis; network-size differences remain unresolved.}
\label{fig:gradients}
\end{figure*}

\textbf{Development-grid contrasts with direct-policy baselines}
(Table~\ref{tab:tally}): \ours{} is resolved positive in all 25 evaluated
non-radius settings against K-EDF, OTM3DQN, RMP-RL-cell, and the idle floor.
The corresponding survival margins are $+0.013$ to $+0.090$ for K-EDF,
$+0.016$ to $+0.099$ for OTM3DQN, and $+0.065$ to $+0.204$ for RMP-RL-cell;
every displayed interval excludes zero.  Reconstructed learned-comparator
intervals use training seed as the unit ($n{=}3$).  The OTM3DQN margin grows
with charger speed.

\textbf{Against \hqarrf{}} (Fig.~\ref{fig:gradients}): the strongest
domain-engineered comparison reveals where the architecture is most favorable:
four settings are resolved positive, while the overall paired difference is not
statistically resolved (mean ${\approx}+0.003$ across settings; 4 resolved
positive, 18 unresolved, 3 resolved negative).  The following patterns are
exploratory and regime-dependent:
\begin{itemize}
\item \emph{Charger capacity} (six levels): monotone, from a resolved positive
  difference at capacity $5{,}000$ ($+0.021$, CI $[+0.006,\allowbreak +0.036]$) through an unresolved difference at the
  center to a resolved negative difference at $20{,}000$ ($-0.034$, CI
  $[-0.047,\allowbreak -0.021]$).
\item \emph{Charging power}: all three perturbed levels have positive point estimates,
  with a resolved positive difference only at power 30
  ($+0.031$, CI $[+0.007,\allowbreak +0.054]$).
\item \emph{Charger speed}: the slowest charger has a resolved negative difference
  ($-0.012$ at $v{=}2.5$); point estimates are positive from $v{=}6.25$
  onward, but those differences remain unresolved.
\item \emph{Consumption}: unresolved through moderate drain, a resolved positive difference at
  $0.005$ ($+0.010$), and a resolved negative difference only at the extreme $0.006$ ($-0.015$).
\item \emph{Network scale}: differences remain unresolved at all five sizes (200--400 nodes).
\end{itemize}

These exploratory patterns trace a coherent boundary: lookahead pays where the
charger's own resources make scheduling the binding problem, and a zero-latency
heuristic suffices where the environment approaches a pure reaction race
(\S\ref{sec:discussion}).  They are not a ranking; the
central-physics sealed bank (\S\ref{sec:confirmatory}) is the primary test.

\section{Sealed Confirmatory Result}\label{sec:confirmatory}

\begin{figure}[t]
\centering
\begin{tikzpicture}
\node[inner sep=0, anchor=south west] (pairedplot) at (0,0)
  {\includegraphics[width=\columnwidth]{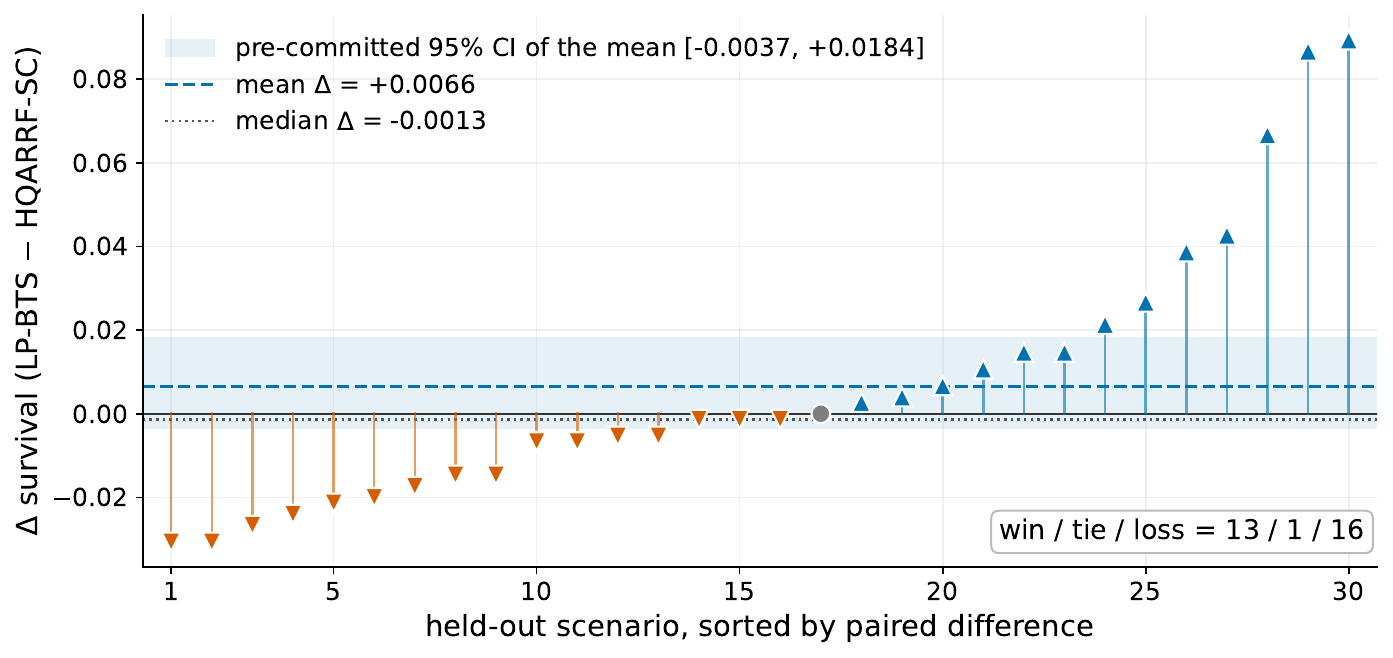}};
\begin{scope}[x={(pairedplot.south east)},y={(pairedplot.north west)}]
  \node[draw=gray!55, fill=white, rounded corners=1pt, inner sep=1pt,
        minimum width=75pt, minimum height=9.5pt, font=\tiny, anchor=south east]
        at (0.992,0.150)
        {R+/U/R$-$ = 13/1/16};
\end{scope}
\end{tikzpicture}
\caption{Sealed-bank paired differences, \ours{} $-$ \hqarrf{}, for the 30
held-out scenarios sorted by difference.  Dashed line: mean; dotted: median;
band: the pre-committed 95\% CI of the mean.  A few large positive
differences and many small negative ones produce a positive mean with a
slightly negative median and a 13/1/16 \rpun{} count.}
\label{fig:paired}
\end{figure}

The evaluation contract --- method identity, contrasts, endpoint, interval
method and interpretation rules --- was committed before the 30 confirmatory
scenarios (generation seeds 600--629, central physics) were generated, and
none of them appears in any training, development or ablation artifact.
Every method was evaluated once on every scenario on a single platform.
Table~\ref{tab:confirmatory} and Fig.~\ref{fig:paired} report the result;
Appendix~\ref{app:repro} records the procedure.

\begin{table*}[t]
\centering\small\setlength{\tabcolsep}{4pt}
\caption{Principal result: sealed one-shot confirmatory bank ($n{=}30$ paired
scenarios, seeds 600--629, central physics, one platform, each scenario
evaluated once).  Learned rows use frozen, trained checkpoints from
three paper-derived training runs.  $\Delta$ = \ours{} $-$ baseline survival at
$T{=}30{,}000$\,s with its 95\% interval and per-unit
\rpun{} counts.
Scenario-unit rows ($n{=}30$) use the pre-committed paired cluster bootstrap;
$^{\dagger}$learned comparators use the training seed as unit ($n{=}3$,
Student-$t$; \S\ref{sec:stats}).}
\label{tab:confirmatory}
\begin{tabular}{lrrrrlr}
\toprule
Method & survival & alive-AUC & travel (m) & $\Delta$ surv. & 95\% CI & R+/U/R$-$ \\
\midrule
\textbf{\ours{} (ours)} & \textbf{0.4545} & \textbf{0.8031} & 53{,}015 & --- & --- & --- \\
\hqarrf{} (primary comparator) & 0.4480 & 0.7990 & 62{,}777 & $+0.0066$ & $[-0.0037,\allowbreak +0.0184]$ & 13/1/16 \\
K-EDF & 0.4048 & 0.7824 & 79{,}011 & $+0.0497$ & $[+0.0427,\allowbreak +0.0567]$ & 30/0/0 \\
OTM3DQN (trained)$^{\dagger}$ & 0.4016 & 0.7783 & 52{,}269 & $+0.0530$ & $[+0.0417,\allowbreak +0.0643]$ & 3/0/0 \\
RMP-RL-cell (trained)$^{\dagger}$ & 0.3316 & 0.7571 & 114{,}933 & $+0.1229$ & $[+0.1169,\allowbreak +0.1288]$ & 3/0/0 \\
NULL (charger idle) & 0.3093 & 0.7478 & 0 & $+0.1452$ & $[+0.1413,\allowbreak +0.1492]$ & 30/0/0 \\
\bottomrule
\end{tabular}
\end{table*}

\textbf{Primary contrast.}  On the sealed bank, \ours{} posts the highest
observed survival ($0.4545$) and alive-AUC ($0.8031$), ahead of \hqarrf{}
($0.4480$, $0.7990$).  The paired survival estimate is $+0.0066$ (median
$-0.0013$; 95\% CI $[-0.0037,\allowbreak +0.0184]$), with a \rpun{} count
of $13/1/16$ ($12/5/13$ at the one-sensor materiality floor).  Under the
interpretation rule fixed before bank access, this is an unresolved difference,
not a superiority claim.  Fig.~\ref{fig:paired} explains the pattern: a few
large \ours{} gains coexist with many small losses, yielding a positive mean
and slightly negative median.

\textbf{Secondary contrasts.}  \ours{} beats every remaining baseline on all
30 scenarios with intervals excluding zero: $+0.050$ over K-EDF and $+0.145$
over the do-nothing floor ($n{=}30$ scenarios), and $+0.053$ over reconstructed
OTM3DQN and $+0.123$ over reconstructed RMP-RL-cell, where the interval
uses training seed as the statistical unit ($n{=}3$; every seed-level contrast
is positive, and every seed is 30/0/0 over scenarios).  Descriptively, \ours{} also travelled
15.5\% less than \hqarrf{} (53{,}015 vs.\ 62{,}777\,m); no movement interval
was pre-specified, so we make no inferential claim about travel.

\textbf{Consistency with development evidence.}  The confirmatory primary
result ($+0.0066$, $[-0.0037,\allowbreak +0.0184]$) sits inside the development central
interval ($+0.0052$, $[-0.0099,\allowbreak +0.0215]$).

\section{Ablations}\label{sec:ablations}

\begin{figure}[t]
\centering
\includegraphics[width=\columnwidth]{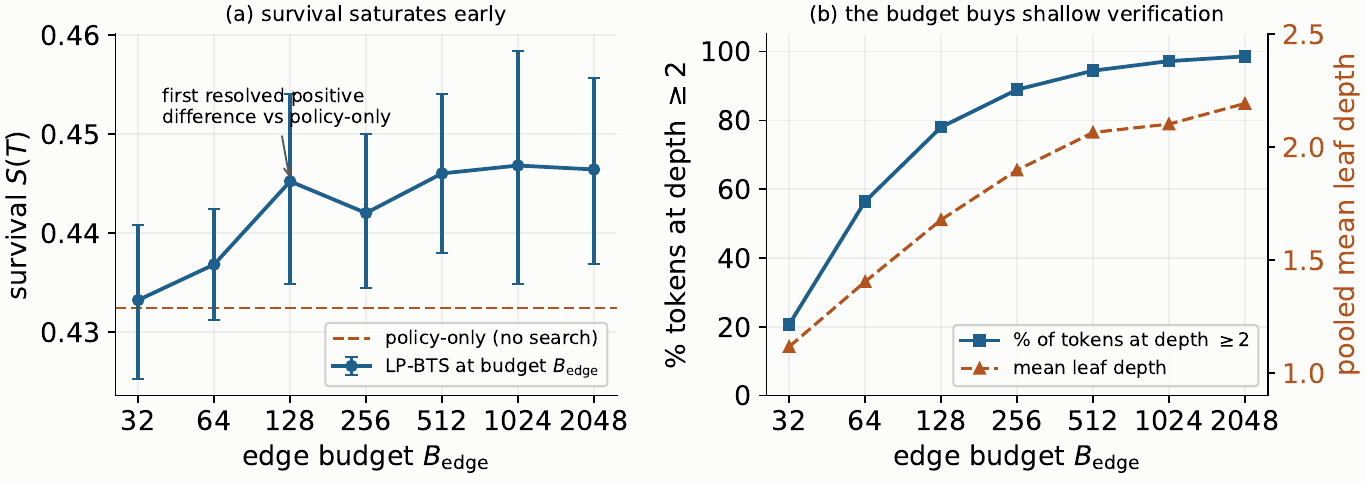}
\caption{Saturation audit (Table~\ref{tab:saturation}); $B_{\mathrm{edge}}$ is
simulated transitions per decision and survival is the 10-world mean.
(a) Survival vs.\ edge budget: whiskers place the paired 95\% CI of each
budget's difference from policy-only around the policy-only baseline.  (b) Fraction of edge
tokens spent below the root and pooled mean leaf depth: the resolved positive difference
at $B_{\mathrm{edge}}{=}128$ coincides with most tokens reaching depth
$\ge 2$, while depth itself stays shallow.}
\label{fig:saturation}
\end{figure}

Four matched-configuration ablations isolate the contributions, all on the
frozen central development scenarios (paired, $n{=}10$).  They were executed
on a separate Linux host; because closed-loop trajectories exhibit small
platform-dependent numerical divergence, every contrast here uses a
same-platform, same-invocation $B_{\mathrm{edge}}{=}2{,}048$ control, and
absolute values are not compared across platforms
(Appendix~\ref{app:repro}):

\begin{itemize}
\item \textbf{Explicit continuation check.}  Holding the learned proposal
  support fixed, replacing PUCT with direct policy selection (the policy
  argmax over the full universe, no tree) sacrifices $0.0140$ survival
  --- about $3.5$ live sensors at $N{=}250$ ---
  (95\% CI $[+0.0044,\allowbreak +0.0232]$, \rpun{} count $6/1/3$) and
  increases travel by 23\% (65{,}094 vs.\ 53{,}036).  Thus search adds a
  measurable continuation check beyond the learned proposal, on both service
  retention and movement.
\item \textbf{Learned prior vs.\ uninformed prior} (uniform-logits stub at
  the same $B_{\mathrm{edge}}{=}2{,}048$, same frozen critic): survival falls
  by $0.0880$ (8.8 percentage points; 95\% CI
  $[+0.0764,\allowbreak +0.0988]$), with a positive paired difference in all
  ten scenarios.  With $K{=}32$ draws from $M\!\approx\!1{,}125$ candidates,
  uniform sampling cannot reliably place valuable stops in the planning
  support.  Its movement was also lower, not higher (40{,}699 vs.\ 53{,}534\,m),
  so the result is not explained by the full method simply travelling farther;
  it is consistent with, but does not prove, better stop targeting by the
  learned proposal.
\item \textbf{Compute budget --- saturation audit} (full method otherwise;
  Fig.~\ref{fig:saturation}, Table~\ref{tab:saturation}), sweeping
  $B_{\mathrm{edge}}\in\{32,\allowbreak 64,\allowbreak 128,\allowbreak
  256,\allowbreak 512,\allowbreak 1024,\allowbreak 2048\}$: explicit search first becomes
  measurably better than policy-only at $B_{\mathrm{edge}}{=}128$
  ($+0.0128$, CI $[+0.0024,\allowbreak +0.0216]$); $B_{\mathrm{edge}}{=}32$ ---
  roughly one simulated transition per sampled arm --- buys nothing
  ($+0.0008$, $[-0.0072,\allowbreak +0.0084]$).  From $B_{\mathrm{edge}}{=}128$ upward
  no budget is measurably better than $B_{\mathrm{edge}}{=}128$ on this
  central development setting, indicating rapid diminishing returns: there
  is no statistically resolved loss relative to the $2{,}048$-edge reference
  at a $16\times$ smaller budget, and at $B_{\mathrm{edge}}{=}512$ ($4\times$
  smaller) mean survival is near-identical ($0.4460$ vs.\ $0.4464$).  This
  characterizes cost against quality on one setting; $B_{\mathrm{edge}}{=}2{,}048$
  remains the frozen configuration.
\item \textbf{Bootstrap vs.\ final} ($A$ vs.\ \ours{}): reported in
  \S\ref{sec:teacher}; statistically unresolved.
\end{itemize}

\textbf{Why shallow search helps.}  Two diagnostics locate the benefit.
First, the improvement begins when computation starts checking continuations,
not merely covering more root arms: at $B_{\mathrm{edge}}{=}64$, about 90\%
of the sampled root arms are already explored (28 of about 31), whereas the
share of tokens at depth ${\ge}2$ rises from 21\% at 32 edges to 78\% at 128
--- precisely where the resolved positive difference appears.  Depth remains
deliberately shallow (pooled mean $1.1\to2.2$, maximum 5): the useful extra
computation is one-step verification of leading stops, not deep rollout.
Second, budgets that match $B_{\mathrm{edge}}{=}2{,}048$ survival do not
reproduce its \emph{decisions}: even $B_{\mathrm{edge}}{=}1{,}024$ chooses the
same initial stop in only 6/10 scenarios yet attains similar survival.  This
is consistent with search resolving continuation effects among several
near-equivalent actions rather than relying on one brittle root choice.

We do not report an ablation swapping one-to-many stops for single-sensor
actions: in this pipeline that swap necessarily changes the physics, the
candidate geometry, the policy inputs and the critic semantics at once, so it
cannot be a single-factor comparison (Appendix~\ref{app:repro}).  One-to-many
charging is therefore presented as the problem formulation, not as an
isolated source of gain.

\section{Computation Budget and Deployment Envelope}\label{sec:cost}

\begin{table}[t]
\centering\small
\caption{Measured per-decision computation (central setting; one CPU core,
no GPU) and the physical decision cadence.}
\label{tab:cost}
\begin{tabular}{lr}
\toprule
Method & per decision \\
\midrule
K-EDF & 0.2\,ms \\
RMP-RL-cell & 3.8\,ms \\
OTM3DQN & 4.8\,ms \\
\hqarrf{} & 35\,ms \\
\ours{} & 41.8\,s (default) \\
\midrule
\multicolumn{2}{l}{\emph{physical cadence (250 dev.\ episodes):}} \\
mean inter-decision interval & 403\,s \\
compute/interval ratio, mean & 0.114 \\
compute/interval ratio, p95 / max & 0.187 / 0.240 \\
\bottomrule
\end{tabular}
\end{table}

The deployment-relevant quantity is the physical decision cadence rather than
a millisecond control loop.  Across 250 development episodes, the mean
simulated interval between decisions is $403$\,s (minimum episode mean
$229$\,s), so the measured compute occupies 11\% on average and at most 24\%
of the window in which the previous action is still executing.  At the frozen
default budget, that compute is ${\sim}41.8$\,s of single-core CPU per decision,
three to five orders of magnitude above the direct baselines
(Table~\ref{tab:cost}).  The implementation remains synchronous (decide, then
act): these measurements motivate, but do not demonstrate, an asynchronous
plan-while-acting realization, and we do not claim real-time deployment.  The
budget is also an explicit design knob.  The saturation audit
(\S\ref{sec:ablations}) finds the survival--latency trade-off flat within CIs
down to $B_{\mathrm{edge}}{=}128$: the frozen configuration is about $4\times$
larger than a near-identical-mean budget and about $16\times$ larger than a
budget with no statistically resolved loss on the central development setting
($n{=}10$).
Because per-decision cost is dominated by simulated transitions (one edge
token each, ${\approx}20$\,ms on the reference platform), latency scales
almost linearly with the budget; Table~\ref{tab:budgetcost} projects the
reference-platform latency from the single measured $B_{\mathrm{edge}}{=}2{,}048$
point via edge count.  Matching survival at $B_{\mathrm{edge}}{=}512$ or
$128$ would cut the per-decision latency from $41.8$\,s to roughly $10$\,s or
$3$\,s respectively --- still far above the direct scheduling baselines
(Table~\ref{tab:cost}), but without a statistically resolved survival loss on
the 10-scenario central development set.  (A budget-independent overhead ---
candidate construction plus one policy evaluation over the initially
${\approx}1{,}125$ stops in the default setting --- adds a small constant,
so the true low-budget latencies are marginally above this proportional
projection.)

\begin{table}[t]
\centering\small
\caption{Per-decision cost vs.\ edge budget on the central setting.  Latency
is projected from the measured $B_{\mathrm{edge}}{=}2{,}048$ point
(Table~\ref{tab:cost}) in proportion to edge tokens; survival deltas are the
same-platform paired audit values (\S\ref{sec:ablations}).  The displayed
recorded-token counter equals $B_{\mathrm{edge}}{+}1$ in the frozen
implementation and is used only for the latency projection; the planning
budget remains $B_{\mathrm{edge}}$ simulated transitions.}
\label{tab:budgetcost}
\resizebox{\columnwidth}{!}{%
\begin{tabular}{rrrl}
\toprule
$B_{\mathrm{edge}}$ & recorded tokens/dec.\ & proj.\ latency & survival vs.\ pol.-only \\
\midrule
128 & 129 & \phantom{0}2.6\,s & $+0.0128$ (resolved +) \\
256 & 257 & \phantom{0}5.2\,s & $+0.0096$ (resolved +) \\
512 & 513 & 10.5\,s & $+0.0136$ (resolved +) \\
1024 & 1025 & 20.9\,s & $+0.0144$ (resolved +) \\
2048 & 2049 & 41.8\,s (meas.) & $+0.0140$ (resolved +) \\
\bottomrule
\end{tabular}}
\end{table}

\section{Discussion}\label{sec:discussion}

\subsection{Learning handles breadth; planning handles continuation}
The ablations expose a clean division of labour.  In an action universe of
about $1{,}125$ stops, learned proposal support is what places promising
actions into a 32-arm planning set; uniform support loses 8.8 survival points.
Once that support is available, shallow search protects roughly 3.5 additional
live sensors at $N{=}250$ and reduces movement relative to direct policy
selection.  The learned critic supplies leaf evaluation, although its
contribution is not independently isolated here.  The search gain appears as
soon as the tree checks one step beyond the root and then flattens; budgets
with comparable survival can still choose different stops.  Thus the method's
value is not a claim of deep search: learning handles breadth, while explicit
short-horizon lookahead resolves continuation effects among near-equivalent
actions.

An independent preregistered learnability audit supports this interpretation.
Across 32 decision states on disjoint training worlds, action ranks under two
continuation policies had Spearman correlation only $+0.0067$; a ridge
diagnostic had held-out ranking correlation $-0.137$, and its best-vs-random
advantage stayed below one sensor at both 1{,}500 and 3{,}000\,s.  Thus action
quality was continuation-dependent in this probe, giving a method-independent
rationale to test explicit lookahead as a decision-making response.  The
short horizons and two continuations do not rule out a learnable direct policy.

\subsection{Planning under a computation budget}
The regime grid delineates an operating envelope rather than claiming uniform
dominance.  Relative to the strongest domain-engineered comparator, LP-BTS is
most favourable where the charger's own resources make scheduling binding
(small battery, high transfer power, fast charger), and less favourable where
the environment approaches a direct reaction race (extreme drain, slow
charger, effectively unconstrained battery).  Together with the budget audit,
this yields a useful design implication: explicit search is most promising
when resource-constrained continuation decisions matter, and much of its
measured value is available at $B_{\mathrm{edge}}{\approx}128$--$512$, well
below the frozen budget.  The sealed bank records the leading observed
central-setting survival and AUC; its difference from the strongest
handcrafted scheduler remains unresolved under the pre-specified rule, while
it is clearly above the deadline and reconstructed learned comparators.

A second envelope is geometric rather than computational.  Because the stop
universe is rebuilt from the live geometry and scored without a fixed output
head, the same frozen checkpoint absorbed a $3.8\times$ change in action-space
size across the evaluated network sizes; a cell-based head cannot follow that
change and coarsens to its fixed lattice instead.  The charging radius pushes
the same distinction further --- it moves a comparator's output width from 471
to 4{,}012 actions and therefore demands a separately trained family per
radius, while \ours{} would need no change (\S\ref{sec:baselines}).  We report
no \ours{} rows on that axis, so this is an architectural property here, not a
measured robustness result; measuring it is the natural next experiment.

\subsection{What may transfer beyond mobile charging}
The division of labor demonstrated here may be relevant when candidate sets are
large and state-dependent, actions are structured, short-horizon simulation is
cheap enough, delayed consequences are not fully captured by direct action
scores, and online computation is limited.  Routing, scheduling, combinatorial
allocation, and agent tool or action selection are possible future settings for
testing this direction.  This study does not establish cross-domain
generalization; those domains were not evaluated here.

\subsection{Limitations}
All results are simulation-only, under one energy model, with a single
charger and a star topology; the charging-radius axis carries no learned
comparator and was therefore not run for \ours{} either
(\S\ref{sec:baselines}), so the reported grid is radius-fixed at $30$\,m.  Development-grid numbers are exploratory and platform-conditioned,
and the held-out confirmatory bank shares the central development physics.  The
per-decision cost remains three to five orders of magnitude above direct
baselines; the budget audit identifies a $16\times$ smaller central-development
setting without statistically resolved survival loss, not a real-time system.
An asynchronous plan-while-acting implementation has not been built, and no
cross-domain validation has been performed.

\section*{Conclusion}
We put learned models inside explicit planning for a large, state-dependent
action space, using one-to-many mobile charging as the concrete instance.  LP-BTS combines a graph proposal policy
that focuses breadth, a learned value critic that evaluates leaves, and
shallow edge-budgeted tree search that checks short simulated futures before
committing an action.  The component evidence is complementary: targeted
proposal support determines which promising stops enter planning, while
explicit search jointly improves retention and reduces travel once that
support is fixed; the search benefit saturates at modest budgets.

On a sealed confirmatory bank evaluated once under a prospectively specified
protocol, LP-BTS achieved the highest observed survival and alive-AUC, with an
unresolved difference from the strongest domain-engineered comparator.  It
exceeded the deadline baseline and the two source-derived direct-policy
reconstructions
on every paired scenario and travelled 15.5\% less than that comparator
descriptively.  The exploratory grid identifies where this architecture is
most favorable.  This study provides controlled evidence about
learning-guided planning in structured dynamic action spaces.


\appendix
\section{Full Per-Setting Results}\label{app:grid}

Table~\ref{tab:grid} records the per-setting exploratory estimates behind the central
comparison.  The following sections give the budget audit, complete statistical
procedure, reproducibility record, and fully specified baseline.

\begin{table*}[t]
\centering\scriptsize\setlength{\tabcolsep}{2.5pt}
\caption{\ours{} vs.\ \hqarrf{}: paired difference on every non-radius setting
($n{=}10$ scenarios each; 95\% paired bootstrap CI).  R+/U/R$-$ denotes
resolved positive / unresolved / resolved negative under the signed one-sensor
materiality rule; exploratory, no multiplicity correction.  The two panels are
one 25-row table.}
\label{tab:grid}
\begin{tabular}{llrrl}
\toprule
Axis & Value & $\Delta$ & 95\% CI & R+/U/R$-$ \\
\midrule
--- & central & $+0.0052$ & $[-0.0099,\allowbreak +0.0215]$ & U \\
capacity & 5000 & $+0.0207$ & $[+0.0057,\allowbreak +0.0363]$ & \textbf{R+} \\
capacity & 7500 & $+0.0135$ & $[+0.0012,\allowbreak +0.0263]$ & \textbf{R+} \\
capacity & 12500 & $-0.0064$ & $[-0.0220,\allowbreak +0.0109]$ & U \\
capacity & 15000 & $-0.0125$ & $[-0.0285,\allowbreak +0.0019]$ & U \\
capacity & 20000 & $-0.0344$ & $[-0.0468,\allowbreak -0.0204]$ & \textbf{R$-$} \\
speed & 2.5 & $-0.0117$ & $[-0.0217,\allowbreak -0.0019]$ & \textbf{R$-$} \\
speed & 3.75 & $-0.0089$ & $[-0.0240,\allowbreak +0.0047]$ & U \\
speed & 6.25 & $+0.0059$ & $[-0.0091,\allowbreak +0.0220]$ & U \\
speed & 7.5 & $+0.0060$ & $[-0.0109,\allowbreak +0.0232]$ & U \\
speed & 8.75 & $+0.0149$ & $[-0.0017,\allowbreak +0.0321]$ & U \\
speed & 10 & $+0.0104$ & $[-0.0063,\allowbreak +0.0297]$ & U \\
power & 20 & $+0.0076$ & $[-0.0156,\allowbreak +0.0299]$ & U \\
\bottomrule
\end{tabular}
\hspace{16pt}
\begin{tabular}{llrrl}
\toprule
Axis & Value & $\Delta$ & 95\% CI & R+/U/R$-$ \\
\midrule
power & 30 & $+0.0309$ & $[+0.0061,\allowbreak +0.0537]$ & \textbf{R+} \\
power & 40 & $+0.0228$ & $[-0.0027,\allowbreak +0.0468]$ & U \\
consumption & 0.003 & $+0.0060$ & $[-0.0104,\allowbreak +0.0208]$ & U \\
consumption & 0.004 & $+0.0072$ & $[-0.0024,\allowbreak +0.0152]$ & U \\
consumption & 0.0045 & $+0.0029$ & $[-0.0092,\allowbreak +0.0157]$ & U \\
consumption & 0.005 & $+0.0100$ & $[+0.0008,\allowbreak +0.0200]$ & \textbf{R+} \\
consumption & 0.0055 & $-0.0031$ & $[-0.0093,\allowbreak +0.0032]$ & U \\
consumption & 0.006 & $-0.0149$ & $[-0.0219,\allowbreak -0.0085]$ & \textbf{R$-$} \\
sensors & 200 & $-0.0037$ & $[-0.0197,\allowbreak +0.0123]$ & U \\
sensors & 300 & $+0.0023$ & $[-0.0111,\allowbreak +0.0157]$ & U \\
sensors & 350 & $+0.0088$ & $[-0.0027,\allowbreak +0.0206]$ & U \\
sensors & 400 & $+0.0025$ & $[-0.0071,\allowbreak +0.0118]$ & U \\
\bottomrule
\end{tabular}
\end{table*}

\section{Search Saturation Audit}\label{app:saturation}
\begin{table}[t]
\centering\scriptsize\setlength{\tabcolsep}{1.75pt}
\caption{Saturation audit on the frozen central setting ($n{=}10$, paired;
reference = same-invocation $B_{\mathrm{edge}}{=}2{,}048$ rerun).
$B_{\mathrm{edge}}{=}0$ is the policy-only arm (no search) and $\Delta$ is the
paired difference against it.  ``tok$\ge$2'' is the fraction of edge tokens at
depth $\ge 2$; ``agr$_0$'' is first-decision agreement with the reference.
Development evidence.}
\label{tab:saturation}
\begin{tabular}{rrrlrrr}
\toprule
$B_{\mathrm{edge}}$ & surv. & $\Delta$ & 95\% CI & depth & tok$\ge$2 & agr$_0$ \\
\midrule
0 & 0.4324 & --- & --- & --- & --- & 2/10 \\
32 & 0.4332 & $+0.0008$ & $[-0.0072,\allowbreak +0.0084]$ & 1.12 & 21\% & 4/10 \\
64 & 0.4368 & $+0.0044$ & $[-0.0012,\allowbreak +0.0100]$ & 1.40 & 56\% & 5/10 \\
128 & 0.4452 & $\bm{+0.0128}$ & $[+0.0024,\allowbreak +0.0216]$ & 1.68 & 78\% & 5/10 \\
256 & 0.4420 & $+0.0096$ & $[+0.0020,\allowbreak +0.0176]$ & 1.90 & 89\% & 6/10 \\
512 & 0.4460 & $+0.0136$ & $[+0.0056,\allowbreak +0.0216]$ & 2.07 & 94\% & 6/10 \\
1024 & 0.4468 & $+0.0144$ & $[+0.0024,\allowbreak +0.0260]$ & 2.10 & 97\% & 6/10 \\
2048 & 0.4464 & $+0.0140$ & $[+0.0044,\allowbreak +0.0232]$ & 2.19 & 99\% & 10/10 \\
\bottomrule
\end{tabular}
\end{table}

Table~\ref{tab:saturation} separates outcome from computation.  Survival and its paired
interval describe the comparison at each budget; depth and token share show
how much computation moves beyond the root; and first-decision agreement
shows whether additional budget changes the chosen action or mainly refines
local evidence.  It is a saturation diagnostic for shallow planning, not a
claim of deep search.

\section{Statistical Procedure: Full Specification}\label{app:stats}
Fixed before any result for \ours{} existed.  (i)~\emph{Units.}  Deterministic
methods and \ours{}: one value per scenario, unit = scenario.  \hqarrf{}: the
three algorithm seeds are averaged within a scenario, unit = scenario.  Learned
baselines: unit = training seed ($n{=}3$); the $3\times$ seed $\times$
scenario cells are three training runs on shared scenarios, not independent
replicates, and are never pooled.  (ii)~\emph{Pairing.}  Differences are
formed per scenario (key: scenario hash within a factor level; scenario index
across levels; (training seed, scenario) for learned methods), then aggregated
over the unit.  (iii)~\emph{Intervals.}  Scenario-unit: 95\% paired cluster
bootstrap, 10{,}000 resamples, RNG seed 20260901.  Seed-unit: the wider of
Student-$t$ ($\mathrm{df}{=}n{-}1$) and a cluster bootstrap over seeds in which
a resampled seed carries its scenarios; the caption names the interval drawn.
(iv)~\emph{Materiality.}  Signed: an improvement is material when the mean
difference exceeds $1/N_0{=}0.004$.  (v)~\emph{Platform.}  Every row within a
statistical comparison comes from one engine on one platform; cross-platform
numerical-stack divergence measurably alters closed-loop trajectories, so
absolute values are never mixed across platforms.  (vi)~\emph{Multiplicity.}
Development setting-level verdicts are reported individually without
correction and labelled exploratory.  The development-grid learned-baseline
intervals in Table~\ref{tab:tally} use the same training-seed hierarchy as the
confirmatory table; the scenario-level legacy intervals are not used in the
manuscript.

\section{Reproducibility and Procedural Record}\label{app:repro}
\textbf{Sealed bank.}  Thirty scenarios were generated from seeds 600--629
with the same construction as the development central family, after the
execution contract (method identity, contrasts, endpoint, interval method,
interpretation rules) had been committed to the repository; none of the 30
scenario hashes appears in any prior artifact.  All methods were run once per
scenario on one host (Linux/aarch64, torch 2.13); there were no infrastructure
failures and no reruns, and no method, checkpoint, baseline or statistical
rule was changed after the results were observed.  Three prewritten
interpretation templates for the abstract and conclusion, corresponding to
predeclared result classes, were committed before the result.

\textbf{Platforms.}  Development-grid rows for \ours{} were produced on Apple
silicon (darwin-arm64, torch~2.8); ablations and the confirmatory bank on
Linux/aarch64 (torch~2.13).  With identical source (matching module hashes),
the global RNG and the recorder flag excluded by controlled reruns, and
within-platform determinism intact, the two platforms' \ours{} survival on the
ten central development scenarios differs by a mean of $0.0008$ (per-scenario
$|\Delta|\le0.016$, sign mixed).  We attribute this to the numerical stack as
a whole and never compare absolute values across platforms.  Per-decision
wall-times measured on the shared development host are load-contaminated, so
the hardware-independent cost axis is edge tokens.

\textbf{Why no atomic-action ablation.}  A matched ablation swapping the canonical one-to-many stop
action for an atomic single-sensor action cannot be a clean single-factor
comparison in this pipeline.  The charging physics recomputes recipients as
\emph{all} live sensors within radius of the stop, so a genuine single-sensor
action would change the environment dynamics; the candidate geometry
(midpoints, circle intersections, enclosing-circle centres) is defined by
multi-sensor coverage; the policy's candidate features are functions of the
recipient set (in-range count, coverage of the energy deficit); and the frozen
critic's value targets were formed under one-to-many futures.  Changing the
action representation therefore necessarily changes the physics, the candidate
universe, the policy inputs, and the critic semantics at once, and no
atomic-trained checkpoint at the frozen $30$\,m geometry exists.  We
consequently present one-to-many charging as the \emph{problem and action
formulation} (as in prior WRSN work~\cite{xie2015,gong2023}) and claim no
independently isolated performance gain from it; the contribution we do
isolate is learning-guided planning over that action space.

\begin{figure}[t]
\centering
\includegraphics[width=\columnwidth]{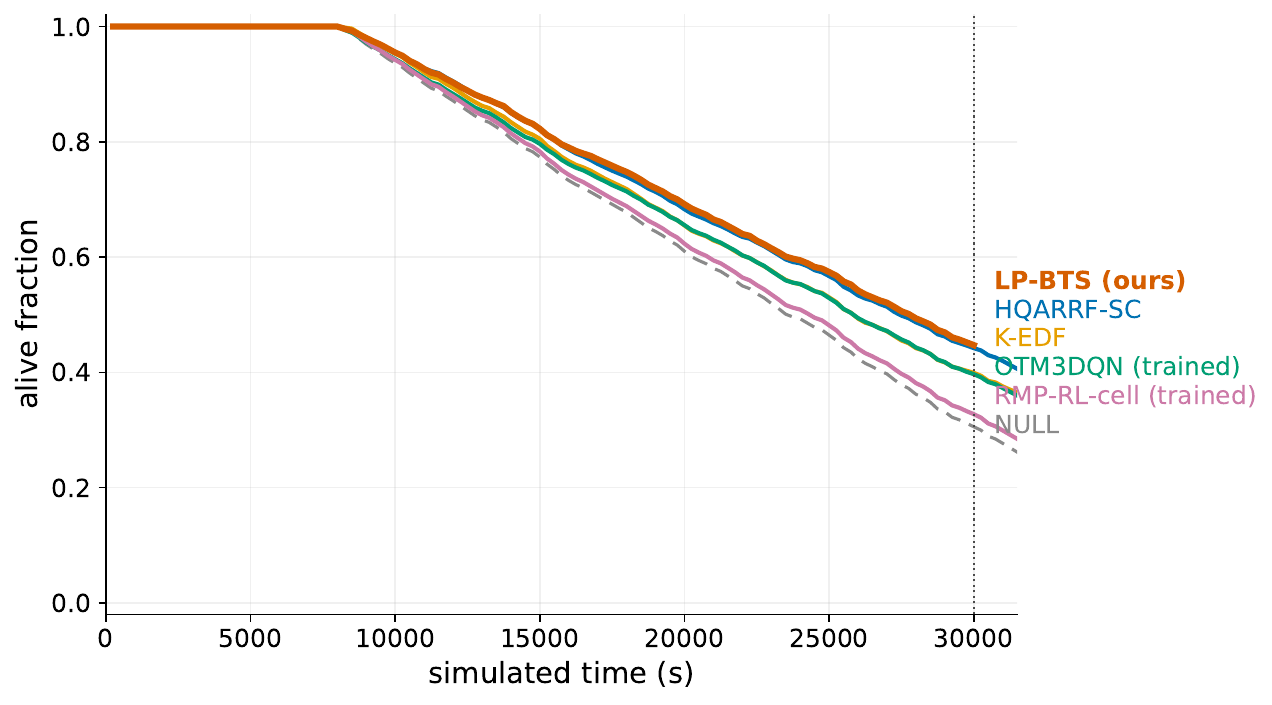}
\caption{Development evidence: alive fraction over time at the central
setting (10-scenario means), displayed through the common
$T{=}30{,}000$\,s evaluation horizon.}
\label{fig:survival}
\end{figure}

\textbf{Development survival curves.}  Fig.~\ref{fig:survival} shows the
alive fraction over time at the central setting for every method.

\textbf{Frozen identities.}  \ours{} = internal checkpoint \ckpt{}: policy
SHA-256 \texttt{0ce2ab97\ldots}, critic \texttt{0153e256\ldots},
$B_{\mathrm{edge}}{=}2{,}048$, $K{=}32$, $c_{\mathrm{puct}}{=}1.5$,
$\tau{=}1$, $\epsilon{=}0.05$, depth ceiling 16, planner seed
$300+\text{decision index}$; learned baselines from the frozen training campaign
(training seeds 101/202/303; trained = 1{,}200 charging rounds).

\normalsize
\section{\hqarrf{} Baseline Specification}\label{app:hqarrf}

{\footnotesize
\hqarrf{} is the single-charger ($M{=}1$) instantiation of the authors'
HQARRF scheduler~\cite{hqarrf}.  Because that work is under review, this
appendix specifies the evaluated variant completely; every value below is the
frozen parameter actually executed (source identity: configuration SHA
\texttt{fb449dc9\ldots}, strategy source SHA \texttt{c90bbc01\ldots}).  No
value was tuned on any bank used in this paper.

\textbf{Hierarchy.}  The field is partitioned into zones by a static
$k$-means clustering of sensor positions (fallback zone count 12, seed 42,
10 initializations).  A tabular Q-learner chooses, every $5$\,s of simulated
time, between \emph{keeping local routing} (action 0) and \emph{requesting an
intervention in zone $z$} (action $z\in\{1,\dots,Z\}$).  The Q-state is the
triple (most-urgent zone, global-risk bucket, imbalance bucket), encoded as
$z\cdot 9 + g\cdot 3 + b$ with three buckets each for $g$ and $b$; the table
has shape $(9Z)\times(1{+}Z)$, is zero-initialized per episode, and is never
carried between scenarios.

\textbf{Risk and urgency.}  Each sensor's risk is a future-time-to-death
estimate with time constant $\tau_{\mathrm{risk}}{=}8{,}000$\,s over the
top-$8$ most urgent nodes, with a missed-deadline bonus of $0.5$; target
ranking uses time-to-death with $\tau_{\mathrm{TTD}}{=}10{,}000$\,s.  Sensors
below $0.42$ of capacity are \emph{critical}, below $0.45$ \emph{requesting};
a global risk above $0.10$ triggers crisis mode.  A soft adaptive threshold
(base $0.36$; consumption, density and under-coverage coefficients
$0.22/0.16/0.18$; weight $3.2$) shifts these cut-offs with local conditions.

\textbf{Local router (ARR-F).}  Within the chosen zone the charger follows an
attraction--repulsion field: electrostatic attraction toward requesting
sensors (energy-deficit weighted, minimum weight $0.1$; softening
$\beta{=}0.2$) and electrostatic repulsion between candidate stops, with a
force gain of $1.6$, interaction radius $220$\,m, at most $16$ interacting
nodes and a $10^{-6}$ numerical floor; the critical fraction for the router
is $0.45$.  A soft task gate (retention $0.999$, override $0.08$) prevents
thrashing between targets.

\textbf{Online tabular update.}  At each decision the previous $(s,a)$ is
updated with $Q(s,a)\leftarrow Q(s,a)+\alpha\,[r+\gamma\max_{a'}Q(s',a')-Q(s,a)]$,
$\alpha{=}0.1$, $\gamma{=}0.9$; $\epsilon$-greedy exploration starts at
$0.3$, decays by $\times0.995$ per update and floors at $0.05$.  The reward
is a weighted sum of observed deltas in alive count ($5$), total risk ($3$),
critical count ($1.5$), delivered energy ($0.5$), deaths ($12$) and movement
($0.005$), minus an intervention cost ($0.1$), plus a zone-risk-gap balance
term ($\lambda{=}2$).  Learning during the evaluated trajectory is the
method's native behaviour, not pre-training; ties and $\epsilon$ draws use a
seeded process RNG (three algorithm seeds, averaged per scenario).

\textbf{Single-charger degeneration.}  With one charger, HQARRF's
charger-assignment, reservation and reassignment machinery is mathematically
trivial (the candidate set has cardinality one) and is left in place
unchanged; the only adaptations are a construction-time guard rejecting
$M{>}1$ and isolation from a foreign partial-charge cap belonging to the
OTM3DQN adapter.  Charging physics, recipient selection, base return and
recharge are the common simulator's, identical for every method.\par
}

\balance

\begin{thebibliography}{16}\footnotesize

\bibitem{kurs2007} A.~Kurs, A.~Karalis, R.~Moffatt, J.~D. Joannopoulos,
P.~Fisher, and M.~Solja\v{c}i\'c, ``Wireless power transfer via strongly
coupled magnetic resonances,'' \emph{Science}, vol.~317, no.~5834, pp.~83--86,
2007.

\bibitem{xie2015} L.~Xie, Y.~Shi, Y.~T. Hou, W.~Lou, H.~D. Sherali, and
S.~F. Midkiff, ``Multi-node wireless energy charging in sensor networks,''
\emph{IEEE/ACM Transactions on Networking}, vol.~23, no.~2, pp.~437--450,
2015.

\bibitem{survey2022} B.~Qureshi, S.~Abdel Aziz, X.~Wang, A.~Hawbani,
S.~H. Alsamhi, T.~Qureshi, and A.~Naji, ``A state-of-the-art survey on
wireless rechargeable sensor networks: perspectives and challenges,''
\emph{Wireless Networks}, vol.~28, no.~7, pp.~3019--3043, 2022.

\bibitem{liu1973} C.~L. Liu and J.~W. Layland, ``Scheduling algorithms for
multiprogramming in a hard-real-time environment,'' \emph{Journal of the ACM},
vol.~20, no.~1, pp.~46--61, 1973.

\bibitem{hqarrf} L.-C. Tao and P.-C. Wang, ``HQARRF: Hierarchical Q-learning
and force-aware routing for multi-charger scheduling in wireless rechargeable
sensor networks,'' manuscript under review, 2026.

\bibitem{cao2021} X.~Cao, W.~Xu, X.~Liu, J.~Peng, and T.~Liu, ``A deep
reinforcement learning-based on-demand charging algorithm for wireless
rechargeable sensor networks,'' \emph{Ad Hoc Networks}, vol.~110, 102278,
2021.

\bibitem{gong2023} Z.~Gong, H.~Wu, Y.~Feng, and N.~Liu, ``Deep reinforcement
learning--based online one-to-multiple charging scheme in wireless
rechargeable sensor network,'' \emph{Sensors}, vol.~23, no.~8, 3903, 2023.

\bibitem{jiang2022} C.~Jiang, Z.~Wang, S.~Chen, J.~Li, H.~Wang, J.~Xiang,
and W.~Xiao, ``Attention-shared multi-agent actor--critic-based deep
reinforcement learning approach for mobile charging dynamic scheduling in
wireless rechargeable sensor networks,'' \emph{Entropy}, vol.~24, no.~7, 965,
2022.

\bibitem{jiang2024} C.~Jiang, W.~Chen, X.~Chen, S.~Zhang, and W.~Xiao, ``Deep
reinforcement learning approach with hybrid action space for mobile charging
in wireless rechargeable sensor networks,'' \emph{Expert Systems with
Applications}, vol.~249, 123752, 2024.

\bibitem{vuong2024} A.~D. Vuong, H.~T. Tran, H.~N.~Q. Pham, Q.~M. Bui,
T.~P. Ngo, and B.~T.~T. Huynh, ``An adaptive charging scheme for large-scale
wireless rechargeable sensor networks inspired by deep Q-network,''
\emph{Neural Computing and Applications}, vol.~36, no.~17,
pp.~10015--10030, 2024.

\bibitem{jnca2025} J.~Li, H.~Wang, S.~Zhang, P.-Y. Kong, and W.~Xiao,
``Reinforcement learning based mobile charging sequence scheduling algorithm
for optimal stochastic event detection in wireless rechargeable sensor
networks,'' \emph{Journal of Network and Computer Applications}, vol.~243,
104301, 2025.

\bibitem{silver2017} D.~Silver et al., ``Mastering the game of Go without
human knowledge,'' \emph{Nature}, vol.~550, no.~7676, pp.~354--359, 2017.

\bibitem{silver2018} D.~Silver et al., ``A general reinforcement learning
algorithm that masters chess, shogi, and Go through self-play,''
\emph{Science}, vol.~362, no.~6419, pp.~1140--1144, 2018.

\bibitem{schrittwieser2020} J.~Schrittwieser et al., ``Mastering Atari, Go,
chess and shogi by planning with a learned model,'' \emph{Nature}, vol.~588,
no.~7839, pp.~604--609, 2020.

\bibitem{danihelka2022} I.~Danihelka, A.~Guez, J.~Schrittwieser, and
D.~Silver, ``Policy improvement by planning with Gumbel,'' in \emph{Proc.
ICLR}, 2022.

\bibitem{rosin2011} C.~D. Rosin, ``Multi-armed bandits with episode context,''
\emph{Annals of Mathematics and Artificial Intelligence}, vol.~61, no.~3,
pp.~203--230, 2011.

\end{thebibliography}
\end{document}